\documentclass[11pt]{article}

\usepackage[preprint]{acl}

\usepackage{times}
\usepackage{xspace}
\usepackage{latexsym}
\usepackage{url}
\usepackage[T1]{fontenc}

\usepackage[utf8]{inputenc}

\usepackage{microtype}

\usepackage{inconsolata}

\usepackage{graphicx}

\usepackage{booktabs}
\usepackage{multirow}
\usepackage{makecell}
\usepackage{colortbl}
\usepackage{xcolor}
\usepackage{amsmath,amssymb}
\usepackage{enumitem}
\usepackage{algorithm}
\usepackage{algorithmic}

\definecolor{caseRed}{RGB}{220,0,0}
\definecolor{planGreen}{RGB}{0,153,102}
\definecolor{codeTeal}{RGB}{0,153,204}
\definecolor{accentMagenta}{RGB}{204,0,153}
\definecolor{noteBlue}{RGB}{0,102,204}
\definecolor{grayText}{gray}{0.25}
\newcommand{\incode}[1]{\texttt{\textcolor{codeTeal}{#1}}}
\newcommand{\keyw}[1]{\textcolor{planGreen}{\texttt{#1}}}
\newcommand{\hilite}[1]{\textcolor{accentMagenta}{#1}}

\newcommand{\Dcal}{\mathcal{D}}
\newcommand{\Qcal}{\mathcal{Q}}
\newcommand{\Tcal}{\mathcal{T}}
\newcommand{\Pcal}{\mathcal{P}}
\newcommand{\Scal}{\mathcal{S}}

\newcommand{\slate}{{\tt    SLATE}\xspace}

\usepackage[most]{tcolorbox}
\lstdefinelanguage{json}{
  basicstyle=\ttfamily\footnotesize,
  showstringspaces=false,
  breaklines=true,
  frame=none
}

\definecolor{promptheader}{RGB}{45,45,45}
\definecolor{promptbg}{RGB}{245,245,245}

\newtcolorbox{promptwrapper}[1]{
  enhanced,
  breakable,
  colback=gray!5,
  colframe=black!60,
  boxrule=0.6pt,
  arc=3pt,
  left=6pt,
  right=6pt,
  top=6pt,
  bottom=6pt,
  title=#1,
  coltitle=white,
  colbacktitle=black!75,
  fonttitle=\bfseries,
  attach boxed title to top left={yshift=-2mm, xshift=3mm},
  boxed title style={
    boxrule=0pt,
    arc=3pt,
    outer arc=3pt
  }
}

\title{Long-Horizon Plan Execution in Large Tool Spaces through Entropy-Guided Branching}

\author{
\textbf{Rongzhe Wei}$^{1,2}$, \textbf{Ge Shi}$^{2}$, \textbf{Min Cheng}$^{2}$, \textbf{Na Zhang}$^{2}$, \\
\textbf{Pan Li}$^{1}$, \textbf{Sarthak Ghosh}$^{2}$, \textbf{Vaibhav Gorde}$^{2}$, \textbf{Leman Akoglu}$^{2}$ \\
$^{1}$Georgia Institute of Technology, $^{2}$Amazon \\
\texttt{\{rongzhew, gegshi, mmcheng, naazhang, ghossart, gordev, lakoglu\}@amazon.com} \\
\texttt{panli@gatech.edu}
}

\begin{document}
\maketitle
\begin{abstract}
Large Language Models (LLMs) have significantly advanced tool-augmented agents, enabling autonomous reasoning via API interactions. However, executing multi-step tasks within massive tool libraries remains challenging due to two critical bottlenecks: (1) the absence of rigorous, plan-level evaluation frameworks and (2) the computational demand of exploring vast decision spaces, stemming from large toolsets and long-horizon planning. To bridge these gaps, we first introduce \slate (\textbf{S}ynthetic \textbf{L}arge-scale \textbf{A}PI \textbf{T}oolkit for \textbf{E}-commerce), a large-scale context-aware benchmark designed for the automated assessment of tool-integrated agents. Unlike static metrics, \slate accommodates diverse yet functionally valid execution trajectories, revealing that current agents struggle with self-correction and search efficiency. Motivated by these findings, we next propose Entropy-Guided Branching (EGB), an uncertainty-aware search algorithm that dynamically expands decision branches where predictive entropy is high. EGB optimizes the exploration-exploitation trade-off, significantly enhancing both task success rates and computational efficiency. Extensive experiments on \slate demonstrate that our dual contribution provides a robust foundation for developing reliable and scalable LLM agents in tool-rich environments. 
\end{abstract}

\section{Introduction}
\label{sec:intro}
\vspace{-2mm}
The rapid progress of LLMs has substantially enhanced the reasoning and decision-making capabilities of AI agents~\citep{wang2024survey,guo2024large,xi2025rise}, enabling them to autonomously interact with external tools and APIs to solve real-world tasks~\citep{yao2023react,shinn2023reflexion}. These tool-augmented LLM agents demonstrate significant potential for automating complex workflows across diverse domains, including e-commerce~\citep{yao2022webshop,yang2024finrobot}, software development~\citep{hong2023metagpt}, scientific discovery~\citep{boiko2023emergent}, and embodied AI~\citep{ahn2024autort}.

Despite this potential, the transition toward context-grounded, long-horizon tool-use planning reveals a significant performance gap. In these sophisticated scenarios, agents are required to navigate massive tool libraries and execute multi-step plans that involve intricate dependencies \citep{toolllm}. We argue that the development of robust tool-use agents is primarily bottlenecked by two intertwined challenges. The first is the \textit{absence of rigorous, plan-level evaluation frameworks} capable of assessing agents in high-complexity settings with diverse trajectories and deterministic outcome rules for scoring \citep{qu2025tool}. The second challenge stems from the \textit{limitations of existing agentic methods in navigating the vast decision spaces} inherent to these tasks, which often results in computational inefficiency and reduced autonomy \citep{huang2023metatool,xu2025llm}.

The first challenge lies in evaluation. Current benchmarks are inadequate, primarily in two respects. Many employ limited toolsets that do not reflect the scale or diversity of real-world applications~\citep{yao2022webshop,yao2024tau,shridhar2020alfworld}. Conversely, benchmarks that do incorporate large tool libraries often assess only single-step tool invocation, neglecting the sequential dependencies crucial for multi-step task execution~\citep{tang2023toolalpaca,gorilla,li2023api}. Furthermore, attempts at plan-level evaluation frequently depend on subjective LLM-as-a-judge assessments~\citep{toolllm,huang2024planning}, which cannot reliably measure an agent's true end-to-end task completion capabilities. This is especially problematic as such methods may penalize valid, alternative solution paths (for instance, those with redundant but harmless tool calls or different but effective execution orders), thereby failing to capture a holistic view of agent proficiency.

\begin{figure*}[t]
  \vspace{-6mm}
  \centering
  \includegraphics[width=0.75\textwidth]{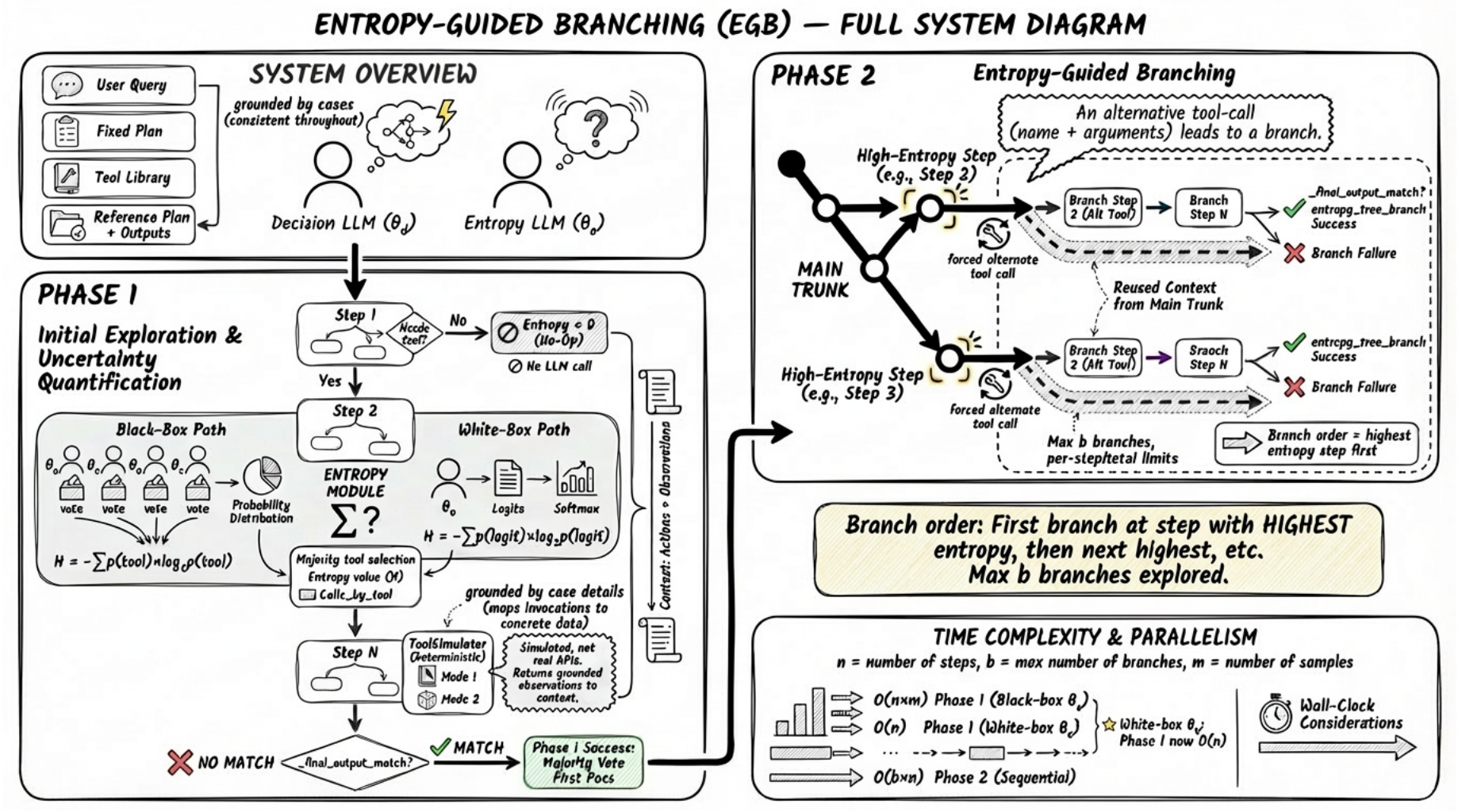}
  \caption{\textbf{Overview of the Entropy-Guided Branching (EGB) framework.} Phase 1 iterates over a fixed plan, using an entropy estimation module (black-box sampling or white-box logit-based) to quantify tool selection uncertainty at each step and execute the majority-voted tool. If the final output does not match the reference, Phase 2 branches at high-entropy steps in descending order, substituting alternative tools from the Phase 1 vote distribution and re-executing downstream steps with updated observations. Tool execution results are produced by a deterministic simulator to enable exact-match success evaluation.}
  \label{fig:egb_overview}
  \vspace{-6pt}
\end{figure*}

Complementary to the evaluation gap, the second challenge pertains to the intrinsic limitations of current agentic algorithms in navigating expansive decision spaces. While foundational methods like ReAct \citep{yao2023react} and self-reflection \citep{shinn2023reflexion, madaan2023self} integrate reasoning with environmental feedback, they often struggle to systematically explore the solution space of long-horizon tasks. To address this, an alternative line of research treats the reasoning process as explicit planning by employing search algorithms to navigate decision trees \citep{koh2024tree,ye2025emergence}. Prominent among these are methods based on Monte Carlo Tree Search (MCTS), which aim to balance exploration and exploitation in decision-making \citep{zhou2023language,murthy2023rex,herr2025llm}. However, the efficacy of these search-based approaches is frequently hindered by prohibitive computational costs, especially in scenarios involving large toolsets and long planning horizons. Consequently, despite ongoing efforts to optimize this trade-off, current methods lack the efficiency and reliability required for practical deployment in complex environments.

We address these intertwined deficiencies by introducing a rigorous evaluation framework and a principled search strategy for tool-augmented agents. We first present \slate (\textbf{S}ynthetic \textbf{L}arge-scale \textbf{A}PI \textbf{T}oolkit for \textbf{E}-commerce), a large-scale, context-aware benchmark for objective plan-level evaluation. Each \slate instance consists of a query, an executable plan, and an associated toolset, supported by context-grounded simulation outputs for valid invocations and explicit default outputs for invalid ones, enabling automated end-to-end evaluation at scale while accommodating trajectory diversity. Leveraging \slate, we conduct a systematic study of representative agent architectures, revealing three key findings: (1) agents struggle to self-correct under binary execution feedback in large action spaces; (2) execution history improves tool and argument selection but has limited influence on the decision of whether to invoke a tool; (3) existing search-based methods, particularly MCTS-based approaches, exhibit an unfavorable utility-to-computation trade-off in long-horizon settings. Motivated by these insights, we propose Entropy-Guided Branching (EGB, see Fig.~\ref{fig:egb_overview}), an uncertainty-aware search algorithm that selectively allocates computation by branching only when predictive entropy over tool choices is high, while following a greedy path under confidence. This adaptive strategy improves the exploration–exploitation balance, reducing unnecessary search overhead and increasing end-to-end task success. Together, \slate, our empirical diagnosis, and EGB establish a more rigorous foundation for reliable LLM agents operating in long-horizon tasks.
\vspace{-1mm}
\section{Related Work}
\vspace{-2mm}
\subsection{Benchmarks and Evaluation Frameworks for Tool Use in LLMs}
\vspace{-1mm}
Existing benchmarks for tool-augmented language models primarily fall into two categories. The first includes small-scale environments such as \textsc{$\tau$-Bench}~\citep{yao2024tau} and \textsc{AlfWorld}~\citep{shridhar2020alfworld}, where the tool or action space is limited to a few dozen predefined options. While these settings support controlled evaluations of decision-making and action selection, they lack the scale and complexity required to assess general-purpose agents operating over realistic tool libraries. The second category comprises benchmarks designed for large-scale tool selection \citep{huang2023metatool,tang2023toolalpaca,gorilla,li2023api,huang2024planning,toolllm}. These benchmarks typically rely either on step-level metrics (e.g., tool awareness, retrieval accuracy) or on proxy signals for plan-level evaluation, most commonly subjective judgments from LLM judges. However, they do not adequately account for end-to-end execution correctness on long-horizon plans, which is essential.

To make this gap explicit, Table~\ref{tab:comprehensive_comparison} compares existing benchmarks along four complementary properties. \textit{Large Tool Spaces} (1,000+ tools) captures step-wise tool-selection difficulty, while \textit{Long-Horizon Execution} (10+ steps) captures plan-level execution complexity. \textit{Step-wise Deterministic Outcomes} indicates whether correct tool-argument invocations produce deterministic outputs that can be reliably propagated to later steps. \textit{Plan-Level Grounded Evaluation} indicates whether each case is grounded with sufficient state information to support deterministic final-state verification. As shown in Table~\ref{tab:comprehensive_comparison}, no prior benchmark satisfies all four properties simultaneously. SLATE is designed to fill this exact gap by combining all four properties. In particular, plan-level grounded evaluation in SLATE checks whether the final execution state matches the ground-truth resolution, even when intermediate tool choices differ from the reference trajectory. This evaluation protocol reflects practical deployments, where multiple APIs can have overlapping functionality and distinct tool sequences can still yield the same correct final outcome.

\begin{table*}[ht]
\centering
\vspace{-2mm}
\caption{Comprehensive Comparison of Agent Tool-Call Datasets.}
\vspace{-2mm}
\label{tab:comprehensive_comparison}
\resizebox{\textwidth}{!}{
\begin{tabular}{lcccc}
\toprule
\textbf{Benchmark} & \makecell{\textbf{Large Tool Spaces} \\ \textbf{(1,000+ Tools)}} & \makecell{\textbf{Long-Horizon Execution} \\ \textbf{(10+ Steps)}} & \makecell{\textbf{Step-wise Deterministic} \\ \textbf{Outcomes}} & \makecell{\textbf{Plan-Level Grounded} \\ \textbf{Evaluation}} \\ 
\midrule
\textbf{SLATE (Ours)} & \textbf{$\checkmark$} & \textbf{$\checkmark$} & \textbf{$\checkmark$} & \textbf{$\checkmark$} \\
TRAJECT-Bench~\citep{trajectbench} & $\checkmark$ & $\checkmark$ & $\checkmark$ & $\times$ \\
Toolathlon~\citep{toolathlon} & $\times$ & $\checkmark$ & $\checkmark$ & $\checkmark$ \\
ToolLLM~\citep{toolllm} & $\checkmark$ & $\times$ & $\checkmark$ & $\times$ \\
Gorilla~\citep{gorilla} & $\checkmark$ & $\times$ & $\checkmark$ & $\times$ \\
UltraTool~\citep{huang2024planning} & $\checkmark$ & $\times$ & $\times$ & $\times$ \\
$\tau$-bench~\citep{taubench} & $\times$ & $\checkmark$ & $\times$ & $\checkmark$ \\
$\tau^2$-Bench \citep{tau2bench} & $\times$ & $\checkmark$ & $\checkmark$ & $\checkmark$ \\
ACEBench~\citep{acebench} & $\checkmark$ & $\times$ & $\checkmark$ & $\times$ \\
HammerBench~\citep{hammerbench} & $\checkmark$ & $\times$ & $\checkmark$ & $\checkmark$ \\
ToolSandbox~\citep{toolsandbox} & $\times$ & $\checkmark$ & $\times$ & $\checkmark$ \\
MetaTool~\citep{huang2023metatool} & $\times$ & $\times$ & $\times$ & $\times$ \\
API-Bank~\citep{apibank} & $\checkmark$ & $\times$ & $\checkmark$ & $\checkmark$ \\
\bottomrule
\end{tabular}
}
\vspace{-2mm}
\end{table*}

\vspace{-1mm}
\subsection{Reasoning and Search in Tool-Augmented Agents}
\vspace{-1mm}
Research on LLM agents has transitioned from eliciting internal reasoning to navigating complex interactions with external tools. Early techniques such as Chain-of-Thought (CoT) \citep{wei2022chain} and Self-Consistency (SC) \citep{wang2022self} established foundational reasoning capabilities. These were later extended via structured search methods such as Tree-of-Thought (ToT) \citep{yao2023tree} and Reasoning via Planning (RAP) \citep{hao2023reasoning} to explore diverse cognitive paths.

A distinct line of research grounds agent decisions in external environments. The ReAct framework \citep{yao2023react} pioneered the interleaving of reasoning with tool execution, a paradigm later refined by mechanisms for self-refinement \citep{madaan2023self} and reflective feedback \citep{shinn2023reflexion} that enable agents to learn from historical errors and environmental observations. As tool libraries scale, tool selection is increasingly framed as a formal search problem. Methods such as LATS \citep{zhou2023language} and REX \citep{murthy2023rex} adapt classical algorithms like Monte Carlo Tree Search (MCTS) to guide agents through multi-step decision spaces. However, the prohibitive computational cost of near-exhaustive exploration in long-horizon tasks remains a critical bottleneck. This tension between search completeness and efficiency motivates our development of Entropy-Guided Branching (EGB), an adaptive and uncertainty-aware search strategy. While Li et al.~\citep{li2026entropygatedbranchingefficienttesttime} also use entropy-gated branching, they study pure test-time reasoning where token-level entropy controls beam expansion during chain-of-thought generation. In contrast, our EGB operates in an interactive tool-use setting: entropy is defined over tool-selection distributions, and branching/backtracking is guided by execution-grounded outcomes at the step level.
\vspace{-1mm}
\section{\texttt{SLATE} Dataset Construction}
\vspace{-1mm}
To address the limitations of existing benchmarks, we introduce \slate by grounding its construction in comprehensive planning-trajectory factual contexts. This section details the development of the \texttt{SLATE} dataset, which is specifically engineered to reflect the complexities of real-world e-commerce scenarios. These environments naturally require long-horizon reasoning across intricate dependencies, such as comparative shopping or inventory-aware checkout processes \citep{yao2022webshop, shridhar2020alfworld}. Furthermore, tasks in this domain typically operate over expansive API libraries that span diverse functionalities ranging from product retrieval to logistics management, which provides a realistic testbed for agent scalability in large decision spaces \citep{toolllm, qu2025tool}. 

\begin{figure}[t]
  \vspace{-2mm}
  \centering
  \includegraphics[width=\columnwidth]{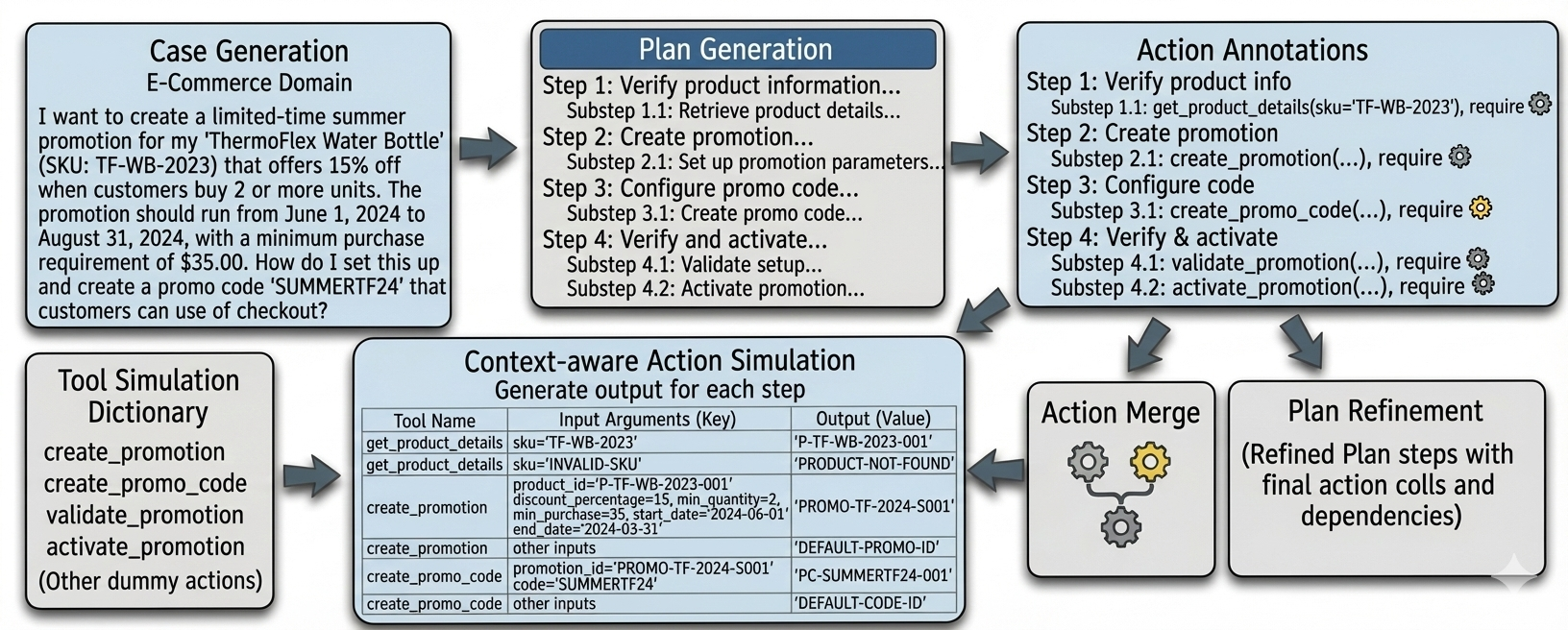}
  \caption{\textbf{Illustration of the \texttt{SLATE} dataset structure.}}
  \label{fig:slate_illustration}
  \vspace{-6pt}
\end{figure}

The resulting dataset comprises user queries paired with multi-step solution plans, precise tool annotations, and simulated execution results for each intermediate step. By providing these grounded execution traces, \texttt{SLATE} facilitates a rigorous assessment of tool-augmented LLM agents, supporting both step-wise verification and end-to-end plan-level evaluation at scale. Our construction pipeline consists of the following stages: query collection, plan generation, tool annotation and normalization, plan refinement, tool simulation, and manual post-processing. This systematic workflow ensures that each trajectory is both logically sound and executionally valid. Detailed prompt designs for each stage are provided in App.~\ref{app:prompt_template}.

\vspace{-1mm}
\subsection{Definition}
\vspace{-1mm}
Let $\Dcal = (\Qcal, \Pcal, \Tcal, \Scal)$ denote the \texttt{SLATE} dataset, where $\Qcal = \{q_1, q_2, \dots, q_n\}$ is a set of $n$ complex user queries, $\Pcal = \{p_1, p_2, \dots, p_n\}$ is the corresponding set of structured, hierarchical plans~\cite{huang2024planning}, $\Tcal = \{t_1, t_2, \dots, t_m\}$ is a library of $m$ tools, and $\Scal$ is a context-aware tool simulator. Each query $q_i \in \Qcal$ is a natural language instruction representing a complex task. Its associated plan $p_i \in \Pcal$ is represented as a sequence of triplets $p_i = [(hs_1, ss_{1,1}, \text{tool}_{1,1}), \dots, (hs_k, ss_{k,l}, \text{tool}_{k,l})]$, where $hs_i$ denotes a high-level step (e.g., ``i''), written in natural language and describing an intermediate subgoal, while $ss_{i,j}$ denotes a finer-grained substep (e.g., ``i,j'') that may invoke a tool calling $\text{tool}_{i,j} \in \Tcal$ or indicate “No Tool Required.” Each tool $t \in \Tcal$ is defined in function-call format as $t(\text{arg}_1, \text{arg}_2, \dots)$, with arguments specified at runtime. Since directly implementing all tools within a large-scale interactive environment is often infeasible, \texttt{SLATE} incorporates a simulator $\Scal$ to approximate the behavior of tool executions. The simulator is designed to generate coherent outputs for each tool call conditioned on the input query and the plan context, thereby enabling consistent and interpretable execution traces. This simulation mechanism plays a crucial role in supporting plan-level evaluation: unlike prior benchmarks that rely on LLM-as-a-judge assessments or isolated step-wise accuracy, our simulator allows objective, automated measurement of end-to-end plan success. Fig.~\ref{fig:slate_illustration} illustrates the dataset structure, and Table \ref{tab:stats} gives summary statistics.

\begin{table}[h]
\centering
\caption{\slate Dataset Statistics}
\vspace{-0.1in}
\resizebox{\columnwidth}{!}{%
\begin{tabular}{lccc}
\toprule
\textbf{Total Tools} & 1000 & \textbf{Relevant Tools in Plans} & 253 \\
\textbf{Avg. Arguments/Tool} & 3.05 & \textbf{Avg. Calls/Tool} & 6.30 \\
\bottomrule
\end{tabular}%
\vspace{-2mm}
\label{tab:stats}
}
\end{table}

\vspace{-1mm}
\subsection{Query Collection}
\vspace{-1mm}
To ensure that the \texttt{SLATE} dataset captures realistic tool-use demands, we focus on domains that inherently require large-scale tool invocation. We select the e-commerce domain as our target setting due to its diverse and interdependent task structure involving frequent interactions between buyers and sellers. This domain naturally necessitates the coordination of numerous APIs and tool calls, making it an ideal testbed for evaluating tool-augmented language agents. 

To balance diversity and topical coherence, we design the query set to cover a broad spectrum of task types while maintaining internal consistency across scenarios. Specifically, we collaborate with domain experts to define 12 representative task categories: \textit{Product Management, Inventory Management, Order Processing, Shipping \& Fulfillment, Pricing \& Promotions, Subscription Management, Customer Service, Returns \& Refunds, Analytics \& Reporting, Catalog Management, Review Management}, and \textit{Miscellaneous}. For each instance, we randomly select a category and employ Claude 3.7 to generate and progressively complicate the query, ensuring both linguistic richness and operational complexity. The prompt template used for generation is provided in Appendix~\ref{app:prompt_template}, with illustrative examples shown in Appendix~\ref{app:examples}. Fig.~\ref{fig:slate_tool_distribution} shows the distribution of tools over the categories.

\begin{figure}[ht]
    \centering
    \includegraphics[width=\linewidth]{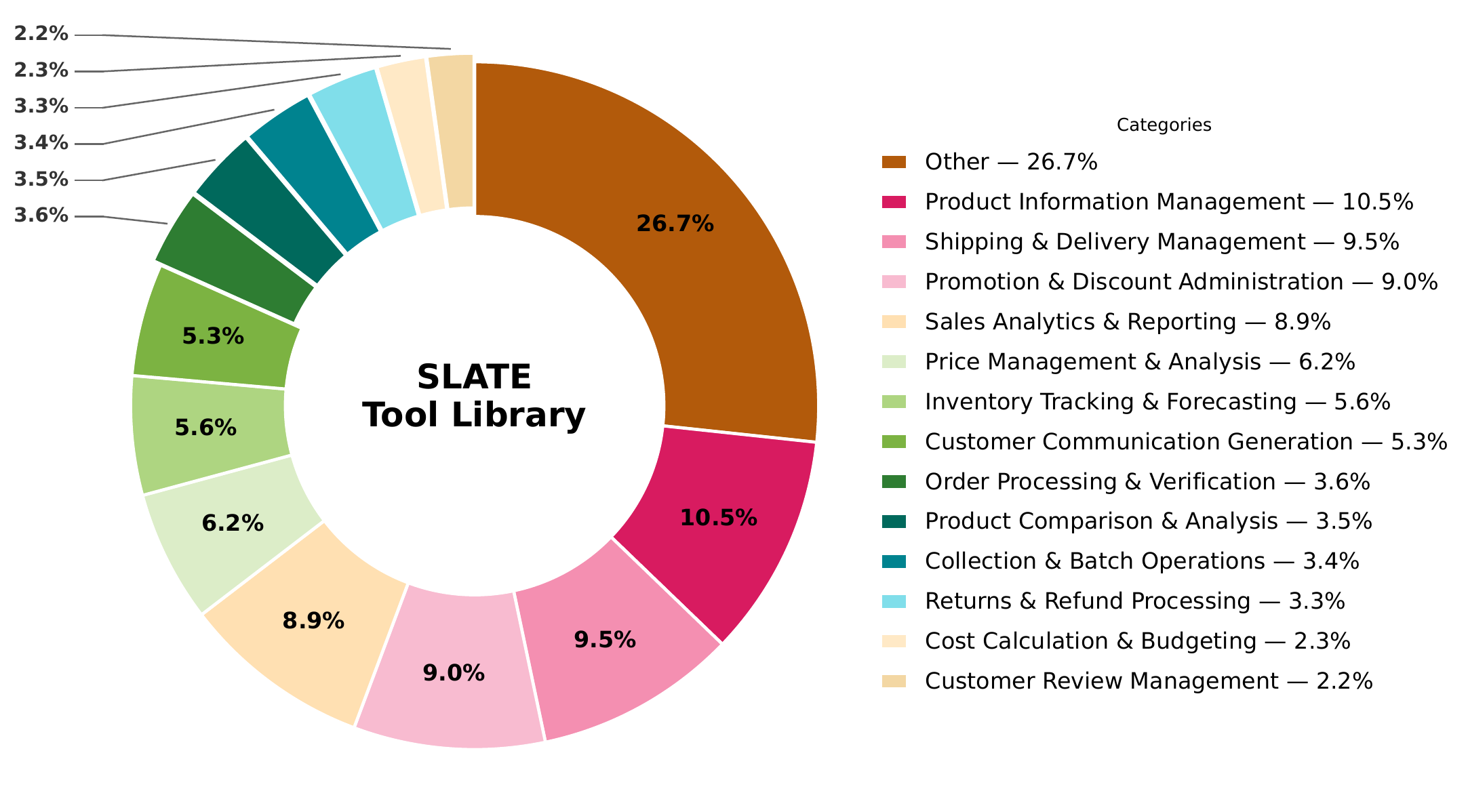}
    \vspace{-5mm}
    \caption{\textbf{SLATE Tool Library category distribution.}}
    \vspace{-2mm}
    \label{fig:slate_tool_distribution}
\end{figure}

\vspace{-1mm}
\subsection{Hierarchical Execution Plans with Tool Invocations}
\vspace{-1mm}
Given a complex natural language query $q \in \Qcal$, we construct a hierarchical execution plan $p \in \Pcal$ in conjunction with a coherent toolkit $\Tcal$, where each substep of the plan is grounded through suitable tool interactions. Our design follows a top-down reasoning paradigm inspired by human problem-solving: the query is recursively decomposed into high-level goals, each of which is further refined into executable substeps. This decomposition enables interpretable multi-stage resolution and  systematic tool grounding aligned with task semantics.

The construction pipeline consists of two key components: (1) Hierarchical Plan Generation, where each query is formulated as an ordered sequence of high-level goals and corresponding substeps; and (2) Tool Definition and Grounding, where tools are identified, created, or reused to support the execution of substeps in a coherent and reusable manner. 

\textbf{Hierarchical Plan Generation.} Given a query $q$, we use Claude 3.7 to generate a multi-step plan  $[(hs_1, ss_{1,1}), (hs_1, ss_{1,2}), \ldots, (hs_k, ss_{k,l})]$, where $hs_i$ denotes a high-level objective, and $ss_{i,j}$ represents a substep that may or may not invoke a tool. Some substeps are purely summarization or decision steps without explicit tool usage. Notably, plan generation is  independent of any toolset to ensure that the resulting structure is free from any tool-specific bias, solely reflects the natural decomposition of the query. 

\textbf{Tool Definition and Grounding.} For each substep requiring tool support, we define a tool in a normalized schema specifying its name, input arguments, functionality, and output format. Several principles guide the tool creation process. First, tool definitions must be specific and not trivially match substep descriptions; overly generic tools or name overlaps with substep text diminish the challenge of tool selection. Second, execution plans often exhibit strong sequential dependencies: the inputs to tools in later steps may rely on the outputs of earlier executions. We explicitly encode such dependencies by ensuring tool arguments are derivable from prior outputs or the original query context. Third, to promote tool reusability and avoid redundancy across queries, we introduce a tool deduplication strategy. Specifically, we embed all generated tool names and descriptions using the ModernBERT~\cite{warner2025smarter}, and for each new query, we embed the query itself and retrieve the top-50 most semantically related tools from the existing library. The LLM is then encouraged to reuse existing tools where appropriate; otherwise, it generates new tools when novel functionality is required. After initial generation, we apply a tool normalization step to ensure consistency across similar tools. Tools are first grouped into functional categories, and within each category, the LLM merges tools with semantically equivalent functionality, reconciling input/output arguments to ensure compatibility. The execution plans are then updated to reflect any merged tool definitions, resulting in a coherent and unified plan-tool structure across the dataset.

\vspace{-1mm}
\subsection{Tool Simulator}
\vspace{-1mm}

Implementing every tool in a massive library is often infeasible. However, tool execution feedback is essential for end-to-end evaluation and providing a closed-loop signal for agentic planning. Since end-users are primarily concerned with final task resolution rather than the specific trajectory of tool calls, we prioritize an evaluation metric centered on execution success. To facilitate this at scale, we develop a Tool Execution Simulator.

The simulator's core function is to provide contextually appropriate outputs while differentiating between correct and incorrect invocations. If a tool call aligns with the ground-truth resolution, the simulator returns a context-coherent result; otherwise, it provides an uninformative response to signal a failed step. Our simulator, along with all other \slate components, is generated using Claude 3.7 and consists of two primary modules:

\textbf{(1) Context-Aware Simulation Database.} This module serves as a repository of ground-truth execution traces. During dataset generation, each execution plan is annotated with the correct sequence of tool calls and their argument dependencies (e.g., \texttt{ToolA(arg1=OUTPUT\_FROM\_STEP\_1)}). Claude 3.7 generates a coherent trace of simulated outputs for each plan, ensuring consistency by conditioning on the query, the full execution plan, and historical input-output pairs. These outcomes are stored in a lookup table structured as \texttt{(tool\_name, arguments) $\rightarrow$ outcome}. For each tool, we also define a default, context-agnostic response for mismatched invocations.

\textbf{(2) Semantic Equivalence Matching.} To handle diverse runtime inputs, the simulator employs a semantic equivalence module to determine if an agent's tool call matches a ground-truth entry. This approach moves beyond simple string comparison, accounting for variations in format (e.g., ``YYYY-MM-DD'' vs. ``MM/DD/YYYY'') and semantic phrasing. If a match is successful, the simulator retrieves the corresponding context-aware outcome; otherwise, it returns the default failure signal. This architecture enables dynamic, realistic simulation for robust end-to-end evaluation.

We remark that \slate and its simulation framework may be of independent interest to the community for research on long-horizon planning and the development of more resilient tool-augmented agents.
\vspace{-1mm}
\section{Preliminaries: Plan-Guided Tool Use}
\vspace{-1mm}
This section formally defines the task of \textit{Plan-Guided Tool Utilization} and reviews the core reasoning and agentic paradigms. 

\vspace{-1mm}
\subsection{Problem Formulation}
\vspace{-1mm}
We formulate the plan-guided tool utilization task as a Markov Decision Process (MDP)~\cite{bellman2015applied}. Given a case query $q \in \mathcal{Q}$, its corresponding plan $p$, and a tool library $\mathcal{T}$, an algorithm $\mathcal{A}$ must sequentially process each substep $ss_{i,j}$ defined in $p$. At each substep $ss_{i,j}$, the algorithm's policy $\pi$ selects an action $a_{i,j}$, which consists of choosing a tool $t_{i,j} \in \mathcal{T} \cup \{\text{NO\_OP}\}$ and generating its corresponding arguments. The policy, $\pi(a_{i,j} | H_{i,j})$, conditions on the execution history $H_{i,j}$, defined as the sequence of outcomes from all preceding substeps: $H_{i,j} = \langle (hs_r, ss_{r,s}, a_{r,s}, o_{r,s}) \rangle_{(r,s) \prec (i,j)}$, where $(r,s) \prec (i,j)$ denotes all index pairs preceding substep $(i,j)$ in the plan's execution order. Each observation $o_{r,s}$ is the result returned by a tool simulator $\mathcal{S}$ upon executing action $a_{r,s}$. Note that pure reasoning-based approaches commit to a sequence of actions without leveraging intermediate observations from the simulator.

\vspace{-1mm}
\subsection{Reasoning and Acting Strategies}
\vspace{-1mm}
Under the MDP framework defined above, agent strategies $\mathcal{A}$ can be broadly categorized into two paradigms. \textit{Reasoning-based} approaches synthesize a complete action sequence $\{a_{i,j}\}$ from a single, self-contained prompt and execute it open-loop, without leveraging intermediate observations $\{o_{i,j}\}$ from the simulator. In contrast, \textit{acting-based} approaches interleave reasoning with interaction, so that decisions are updated online using execution feedback $\{o_{i,j}\}$.

Reasoning-based approaches therefore decouple planning from execution: actions are fixed before interaction, which simplifies rollout but limits in-trajectory correction.

Acting-based approaches treat task resolution as interactive, where the policy $\pi(a_{i,j} \mid H_{i,j})$ is continuously informed by feedback $\{o_{i,j}\}$. The foundational method, \textit{ReAct}~\cite{yao2023react}, establishes an iterative cycle of \textit{Thought-Action-Observation}, using feedback from the simulator $\mathcal{S}$ to inform the next step. \textit{Reflexion}~\cite{shinn2023reflexion} enhances this with a self-correction mechanism, where case-level failures trigger reflection to generate guiding principles for subsequent attempts. More sophisticated strategies employ lookahead search over the state-action space. For instance, MCTS-based methods like \textit{LATS}~\cite{zhou2023language} build a search tree using simulator-driven rollouts, enabling the agent to evaluate longer-term action consequences before commitment.
\vspace{-1mm}
\section{EGB for Long-Horizon Planning}
\label{sec.methodology}
\vspace{-1mm}

\begin{figure}[t]
    \centering
    \includegraphics[width=\columnwidth]{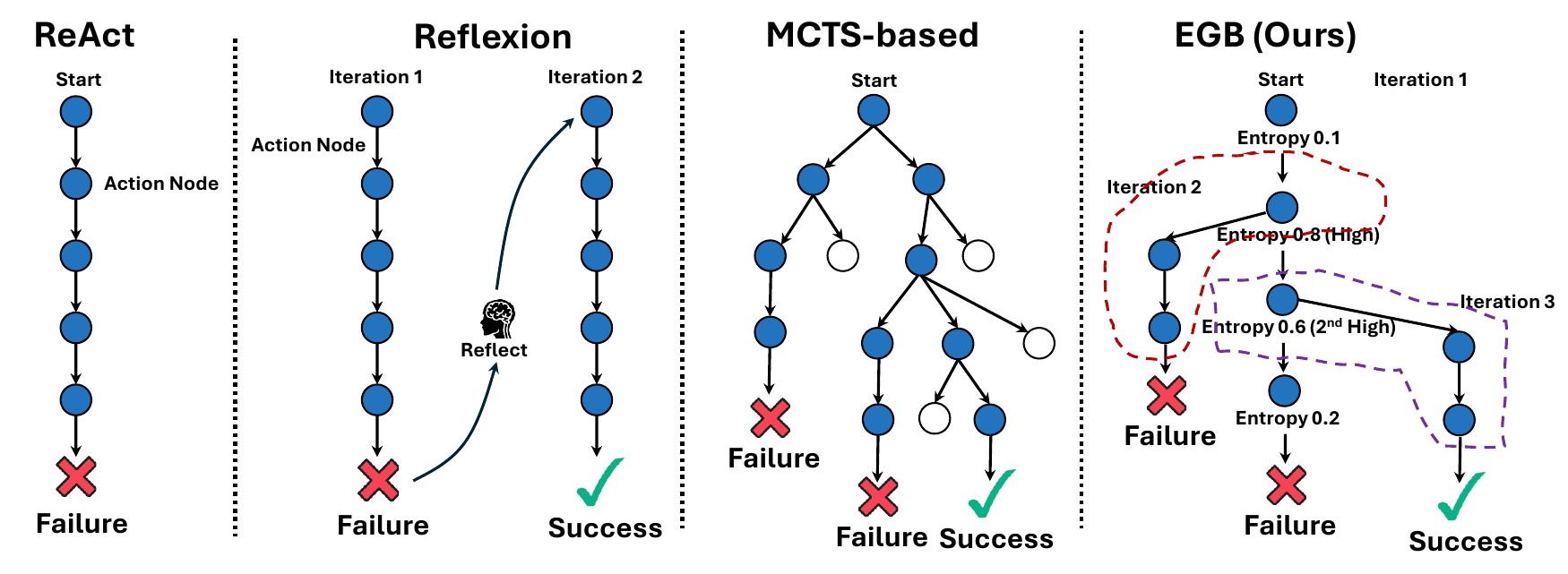} 
    \vspace{-7mm}
    \caption{Comparison of different paradigms.}
    \vspace{-3mm}
    \label{fig:method_comparison}
\end{figure}

We propose \textit{Entropy-Guided Branching (EGB)} (illustrated in Fig.~\ref{fig:method_comparison}), an uncertainty-aware acting strategy designed to navigate expansive tool spaces through iterative trajectory refinement in black-box settings. Unlike pure reasoning-based approaches, EGB operates within a MDP framework, where the agent's policy $\pi(a_{i,j} | H_{i,j})$ is continuously informed by environmental feedback $o_{i,j}$ from a tool simulator $\mathcal{S}$.

\subsection{Initial Exploration and Uncertainty Quantification}
In the first iteration, EGB performs a sequential execution of the plan $p$ inspired by the ReAct paradigm while evaluating the uncertainty of every substep. For each substep $ss_{i,j} \in p$, we assume a black-box setting where internal model logits are inaccessible. To quantify decision uncertainty, we generate $m$ candidate actions $\{a_{i,j}^{(1)}, a_{i,j}^{(2)}, \dots, a_{i,j}^{(m)}\}$ through independent samplings from the policy:
\begin{equation}
a_{i,j}^{(1 \dots m)} \sim \pi(a_{i,j} | H_{i,j})
\end{equation}

The agent executes the final action $a_{i,j}$ determined by a majority vote among these $m$ candidates and receives the observation $o_{i,j} = \mathcal{S}(a_{i,j})$. Simultaneously, we calculate the predictive entropy $E_{i,j}$ based on the distribution of the $m$ samples to serve as a proxy for uncertainty.  Importantly, this estimation method relies solely on output diversity and does not require access to internal model logits or hidden states. While EGB can utilize internal signals if available, as demonstrated in our experiments, this sampling-based approach ensures that our strategy remains fully compatible with black-box proprietary models. The initial execution continues until the agent either reaches a terminal state or the plan fails to achieve the goal.

\subsection{Ranked Branching and Iterative Search}
If the initial trajectory results in failure, EGB leverages the recorded entropy signals to optimize the exploration-exploitation trade-off. We rank all substeps $ss_{i,j}$ of the failed trace in descending order of their entropy values $\{E_{i,j}\}$. This ranking identifies decision points where the model was least confident (highest entropy), suggesting that the optimal tool might be a plausible alternative rather than the top-voted choice. In subsequent iterations, EGB initiates branching from the step with the highest remaining uncertainty. Specifically, it selects a previously unexecuted candidate action from the original sampling set $\{a_{i,j}^{(1 \dots m)}\}$ to spawn a new trajectory. The agent then resumes execution from this new state by following the plan-conditioned history until a successful terminal state is identified or the branching budget is exhausted.

In essence, unlike MCTS-based methods that construct full search trees, EGB performs sequential replays with selective branching triggered at high-uncertainty decision points along failed trajectories. This design offers key advantages in tool-rich environments: while MCTS branches based on exploration-exploitation scores that become noisy in long-horizon tasks, EGB uses entropy as a localized diagnostic signal. Moreover, MCTS typically branches during rollouts before observing outcomes, whereas EGB explicitly leverages failed trajectory information to pinpoint and rectify the specific decisions responsible for failure, avoiding prohibitive exhaustive searches in large action spaces.
\vspace{-1mm}
\section{Experiments}
\label{sec:exp}
\vspace{-1mm}

We set up experiments on the \slate synthetic dataset to evaluate EGB against representative baselines, and conduct studies on the impact of hyperparameter and computational costs.

\begin{table}[ht]
\vspace*{-0.05in}
\caption{Evaluation of search methods on the \slate synthetic dataset with Claude-Sonnet-4.}
\vspace{-0.1in}
\centering
\resizebox{\linewidth}{!}{  
\begin{tabular}{c c c c}
\toprule
\multirow{4}{*}{\makecell{\textbf{Method}}} & \multicolumn{1}{c}{\textbf{Plan-level}} & \multicolumn{2}{c}{\textbf{Step-wise}} \\
\cmidrule(lr){2-2} \cmidrule(lr){3-4}
 & \makecell{Execution\\Success\\Rate} & \makecell{Tool\\Match\\Rate} & \makecell{Action\\Identification\\Accuracy} \\
\midrule
\makecell{\textbf{Baseline-LLM}} & $\backslash$ & 65.2 $\pm$ 0.8 & $\backslash$ \\
\makecell{\textbf{ReAct}} & 29.3 $\pm$ 1.2 & 66.4 $\pm$ 0.4 & 85.7 $\pm$ 0.1 \\
\makecell{\textbf{Reflexion}} & 44.7 $\pm$ 2.3 & 61.5 $\pm$ 1.6 & 83.4 $\pm$ 0.5 \\
\makecell{\textbf{LATS}} & 36.5 $\pm$ 2.5 & 63.4 $\pm$ 1.3 & 87.3 $\pm$ 0.5 \\
\midrule
\makecell{\textbf{EGB-Sampling (Ours)}} & \textbf{54.0 $\pm$ 2.0} & 68.5 $\pm$ 1.5 & 87.9 $\pm$ 0.2 \\
\bottomrule
\end{tabular}
}
\label{tab:method_performance_slate_claude}
\end{table}

\begin{table}[ht]
\vspace*{-0.1in}
\caption{Evaluation of search methods on the \slate synthetic dataset with Qwen2.5-7B-Instruct.}
\vspace{-0.1in}
\centering
\resizebox{\linewidth}{!}{  
\begin{tabular}{c c c c}
\toprule
\multirow{4}{*}{\makecell{\textbf{Method}}} & \multicolumn{1}{c}{\textbf{Plan-level}} & \multicolumn{2}{c}{\textbf{Step-wise}} \\
\cmidrule(lr){2-2} \cmidrule(lr){3-4}
 & \makecell{Execution\\Success\\Rate} & \makecell{Tool\\Match\\Rate} & \makecell{Action\\Identification\\Accuracy} \\
\midrule
\makecell{\textbf{Baseline-LLM}} & $\backslash$ & 23.2 $\pm$ 0.7 & $\backslash$ \\
\makecell{\textbf{ReAct}} & 29.3 $\pm$ 1.2 & 30.0 $\pm$ 0.5 & 78.2 $\pm$ 0.4 \\
\makecell{\textbf{Reflexion}} & 44.2 $\pm$ 2.1 & 25.2 $\pm$ 0.7 & 82.9 $\pm$ 2.9 \\
\midrule
\makecell{\textbf{EGB-Sampling (Ours)}} & \textbf{51.3 $\pm$ 6.0} & 33.6 $\pm$ 2.4 & 83.0 $\pm$ 0.6 \\
\makecell{\textbf{EGB-Logits (Ours)}} & \textbf{67.8 $\pm$ 4.5} & 36.4 $\pm$ 0.8 & 85.1 $\pm$ 0.4 \\
\bottomrule
\end{tabular}
}
\label{tab:method_performance_slate_qwen}
\end{table}

\begin{table*}[ht]
\caption{Comparison of computational cost per case across different search methods on SLATE synthetic dataset.}
\vspace{-0.1in}
\centering
\resizebox{\textwidth}{!}{  
\begin{tabular}{l c c c c c}
\toprule
& \textbf{Claude-Sonnet-4} & \multicolumn{4}{c}{\textbf{Qwen2.5-7B-Instruct}} \\
\cmidrule(lr){2-2} \cmidrule(lr){3-6}
\textbf{Method} & \makecell{\textbf{Running Time}\\\textbf{(seconds)}} & \makecell{\textbf{Running Time $^*$}\\\textbf{(seconds)}} & \makecell{\textbf{Input Tokens}} & \makecell{\textbf{Output Tokens}} & \makecell{\textbf{LLM Invokes}} \\
\midrule
\textbf{Baseline-LLM} & 77 & 138 & $1.83 \times 10^5$ & $3.8 \times 10^3$ & 32.0 \\
\textbf{ReAct} & 195 & 150 & $1.95 \times 10^5$ & $4.0 \times 10^3$ & 32.0 \\
\textbf{Reflexion} & 766 & 707 & $8.90 \times 10^5$ & $2.41 \times 10^4$ & 122.2 \\
\textbf{LATS} & 620 & $\backslash$ & $\backslash$ & $\backslash$ & $\backslash$ \\
\textbf{EGB-Logits (Ours)} & $\backslash$ & 254 & $(1.61+4.86) \times 10^5$ $^{\dagger}$ & $6.4 \times 10^3$ & 44.6 + 149.6 $^{\dagger}$ \\
\textbf{EGB-Sampling (Ours)} & 718 & 1,029 & $2.19 \times 10^6$ & $3.17 \times 10^4$ & 176.4 \\
\bottomrule
\end{tabular}
}
\raggedright
\footnotesize
$\dagger$ For EGB-Logits, input token usage consists of $1.61 \times 10^5$ from generation queries and $4.86 \times 10^5$ from logits-only forward passes (no output tokens). LLM invokes includes 44.6 generation calls and 149.6 lightweight forward passes. $^*$ Parallelization across all methods are disabled for Qwen experiments.
\label{tab:cost_comparison}
\vspace{-3mm}
\end{table*}
\vspace{-0.2in}

\subsection{Experimental Settings} 

\textbf{Baselines:} We evaluate EGB against both reasoning-based and acting-based methods. 
For the former, we consider \textbf{Baseline-LLM}, which selects tools without access to execution history or trajectory-level reasoning and is therefore excluded from plan-level evaluation. For acting-based methods, we compare against: \textbf{ReAct}, which interleaves reasoning traces with tool execution in a sequential manner; \textbf{Reflexion}, which incorporates self-reflection by analyzing execution feedback to refine subsequent actions; and \textbf{LATS}, which builds an entire search tree with environment-aware rollouts.

\textbf{Model:} We conduct primary experiments on Claude-4-Sonnet, a proprietary black-box model with inaccessible logits, to demonstrate that EGB is applicable in real-world deployment settings. To validate generalizability, we additionally evaluate on Qwen2.5-7B-Instruct~\citep{qwen2025qwen25technicalreport}, an open-source model with accessible logits. We selected Qwen2.5-7B based on preliminary experiments indicating competitive instruction-following performance and stable entropy-uncertainty correlation among models of comparable size. For cross-benchmark validation, we also use GPT-OSS-120b~\citep{openai2025gptoss120bgptoss20bmodel}, an open-weight reasoning model with an efficient mixture-of-experts transformer architecture. We use AWS Bedrock to access Claude and GPT models, and $5$ NVIDIA L4 GPUs for Qwen experiments.

\textbf{Evaluation Metrics:} We assess performance from two complementary perspectives: \textbf{plan-level} and \textbf{step-wise} evaluation. 

\begin{itemize}[leftmargin=2mm, itemsep=1pt]
\vspace*{-0.05in}
    \item \textit{Step-wise evaluation} analyzes decision-making at each individual step through two metrics: Action Identification Accuracy measures the proportion of steps where the agent correctly determines whether a tool call is needed. Tool Match Rate evaluates, among steps requiring tools, the alignment rate between the agent's tool selection and the reference tool in the plan.

    \item \textit{Plan-level evaluation} assesses the correctness of the final execution result. We report Execution Success Rate, defined as the proportion of cases where the final execution results exactly match the annotated reference resolutions in the plan, regardless of intermediate missteps that do not affect the outcome. For methods with plan-level retries (Reflexion, EGB, and LATS), this is equivalent to \textit{pass@k} as introduced in $\tau$-bench~\citep{taubench}, i.e., the probability of reaching a success before exhausting $k$ trials. For EGB, we denote this budget by $b$.

\end{itemize}

\textbf{Configurable Hyperparameters:} EGB introduces two critical hyperparameters: \(m\), the number of samplings used to estimate the predictive entropy of tool selection at each step, and \(b\), the branching budget that limits the number of iterations for replanning and exploration. Unless otherwise specified, we set \(m=10\) and \(b=5\) in experiments. For all experiments, we set LLM inference hyperparameters temperature = 1. These settings ensure consistent and comparable results across all baseline methods and model configurations.

\vspace{-1mm}
\subsection{Experimental Results}
\vspace{-1mm}

\textbf{Results on a Proprietary Model:} We first conduct comprehensive experiments on Claude-4-Sonnet to evaluate EGB against baseline methods, with results presented in Table~\ref{tab:method_performance_slate_claude}. At the plan level, EGB achieves substantial improvements in execution success rate, outperforming ReAct by 24.7\% and Reflexion by 9.3\%. Notably, EGB also surpasses LATS by 17.5\% despite LATS being constrained to 25 global search node visits per case---an empirical limit chosen to match EGB's average computational budget. This significant plan-level advantage, coupled with only marginal improvements in step-wise metrics (2.0\%--7.0\% on Tool Match Rate), reveals a critical limitation of baseline methods: without access to step-level correctness feedback from the simulator, they struggle to identify error-prone decisions under constrained computation. Unlike EGB, which leverages internal policy uncertainty (entropy) to pinpoint likely failure points, baseline methods rely solely on end-of-trajectory feedback. In long-horizon tasks, this leaves them unable to effectively localize which steps in the Markov decision process are responsible for failure. This effect is particularly evident for Reflexion and LATS, which exhibit \textit{decreased} Tool Match Rates compared to Baseline-LLM and ReAct, suggesting they modify previously correct tool selections into incorrect ones during their search process. Their plan-level improvements thus stem primarily from exploring more execution paths rather than intelligently targeting problematic decisions, resulting in occasional successes by chance rather than systematic error correction.

\begin{table}[t]
\centering
\vspace{-0.05in}
\caption{Entropy-error relationship on \slate using EGB-Logits.}
\vspace{-0.1in}
\resizebox{0.9\linewidth}{!}{
\begin{tabular}{lccc}
\toprule
\textbf{Entropy Range} & \textbf{\#Steps} & \textbf{\#Errors} & \textbf{Error Rate} \\
\midrule
$[0.000, 0.640)$ & 365 & 184 & 50.7\% \\
$[0.640, 1.281)$ & 116 & 79 & 68.1\% \\
$[1.281, 1.921)$ & 67 & 52 & 77.6\% \\
$[1.921, 2.561)$ & 21 & 20 & 95.2\% \\
$[2.561, 3.202]$ & 15 & 14 & 93.3\% \\
\bottomrule
\end{tabular}
}
\label{tab:entropy_error_slate}
\vspace{-2mm}
\end{table}

\textbf{Results on an Open-source Model:} To further validate our approach, we conduct experiments on Qwen2.5-7B-Instruct, a white-box model that enables direct computation of entropy from output logits rather than sampling-based estimation. As shown in Table~\ref{tab:method_performance_slate_qwen}, the findings mirror those observed with Claude-Sonnet-4: EGB substantially outperforms baseline methods at the plan level while achieving marginally better step-wise performance. We omit LATS from this evaluation due to the prohibitive computational cost of hosting open-source models for extensive tree search. Notably, EGB-Logits achieves 67.8\% execution success rate, surpassing EGB-Sampling by 16.5\%. This observation reinforces our core assumption that entropy serves as a reliable indicator of error-prone decisions. Besides, accurate entropy computation from logits proves more effective than sampling-based estimation.

\textbf{Error Rate and Entropy Correlation on \slate:} To assess whether entropy provides a useful branching signal, we analyze EGB-Logits at the step level, where error is defined as a first-pass tool mismatch against the reference trajectory. As shown in Table~\ref{tab:entropy_error_slate}, correct steps have lower entropy than error steps (mean $=0.052$ vs. $0.119$), and error rate rises from 50.7\% to 93.3\% across entropy bins (with a small fluctuation in the highest-entropy bins). These results indicate that entropy is an effective relative prioritization signal for branch selection rather than a strict binary error detector. In successful branch corrections, errors are reduced through direct tool fixes, cascade fixes via corrected execution context, and functionally equivalent alternative tools; detailed breakdowns are provided in Appendix Sec.~\ref{sec:app_error_entropy}.

\begin{figure}[t]
    \centering
    \vspace{-2mm}
    \includegraphics[width=\linewidth]{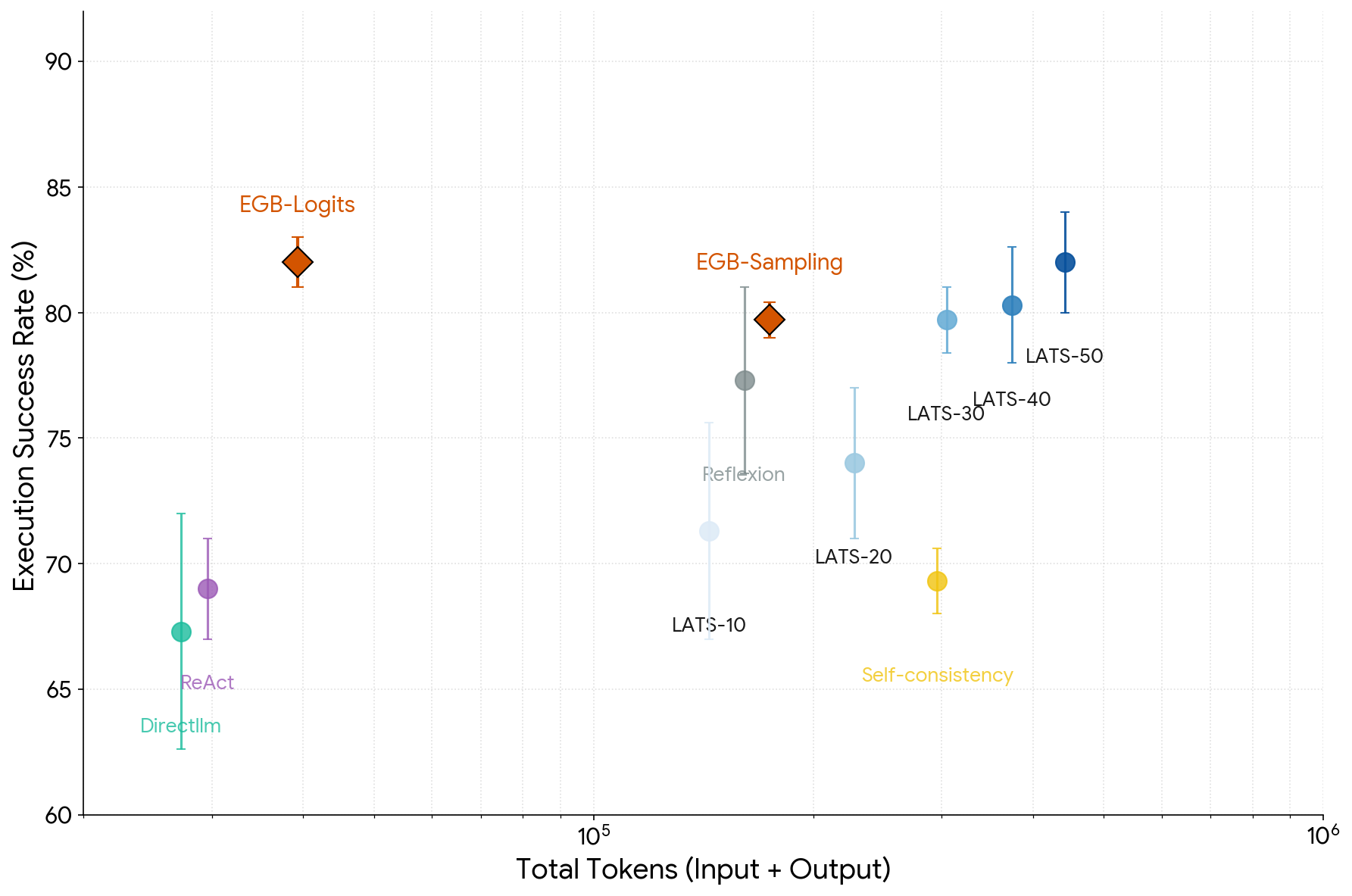}
    \vspace{-5mm}
    \caption{Cross-benchmark validation on adapted UltraTool with GPT-OSS-120b.}
    \label{fig:ultra_tool}
\vspace{-2mm}
\end{figure}

\textbf{Cross-Benchmark Validation on Adapted UltraTool:} Because \slate is synthesized with Claude-family models, we additionally evaluate on adapted UltraTool to reduce potential model-family bias and test whether our conclusions transfer beyond \slate. We choose UltraTool because its plans are shorter-horizon, enabling practical comparison against higher-budget tree-search baselines that are too expensive on \slate. In Figure~\ref{fig:ultra_tool} (details in Appendix Sec.~\ref{sec:app_ultratool}), \textbf{Self-Consistency} samples multiple ($m=10$) tool choices per step and executes the majority-voted action, while \textbf{LATS-$k$} denotes LATS with an upper limit of $[10,50]$ distinct search-tree nodes visited. EGB remains consistently strong: EGB-Logits matches the best reported success rate (LATS-50) while using only a fraction of the token cost, and EGB-Sampling also achieves high performance comparable to LATS-30; meanwhile, LATS shows the expected budget--accuracy trade-off, with success improving as node budget increases but with diminishing returns.

\textbf{Computation Cost:} Table~\ref{tab:cost_comparison} presents the computational cost comparison. For Qwen experiments, we disable parallelization across all methods to ensure fair comparison on a single GPU, resulting in EGB-Sampling being slower than Reflexion (1,029s vs. 707s). However, the $m$ independent samples in EGB-Sampling can be trivially parallelized---as demonstrated in Claude experiments with 2 concurrent workers, where EGB-Sampling achieves 718 seconds (\textbf{6.3\%} faster than Reflexion). Compared with LATS (620s on Claude), EGB-Sampling incurs moderate extra runtime but delivers substantially higher execution success in Table~\ref{tab:method_performance_slate_claude}. EGB-Logits demonstrates the strongest efficiency, requiring only 254 seconds (\textbf{2.8$\times$} faster than Reflexion) by leveraging lightweight forward passes without autoregressive generation. Notably, EGB-Logits reduces output tokens by \textbf{73\%} compared to Reflexion ($6.4 \times 10^3$ vs. $2.41 \times 10^4$). These results confirm that entropy-guided branching achieves strong performance without prohibitive computational overhead.

\begin{figure}[t]
    \centering
    \includegraphics[width=\linewidth]{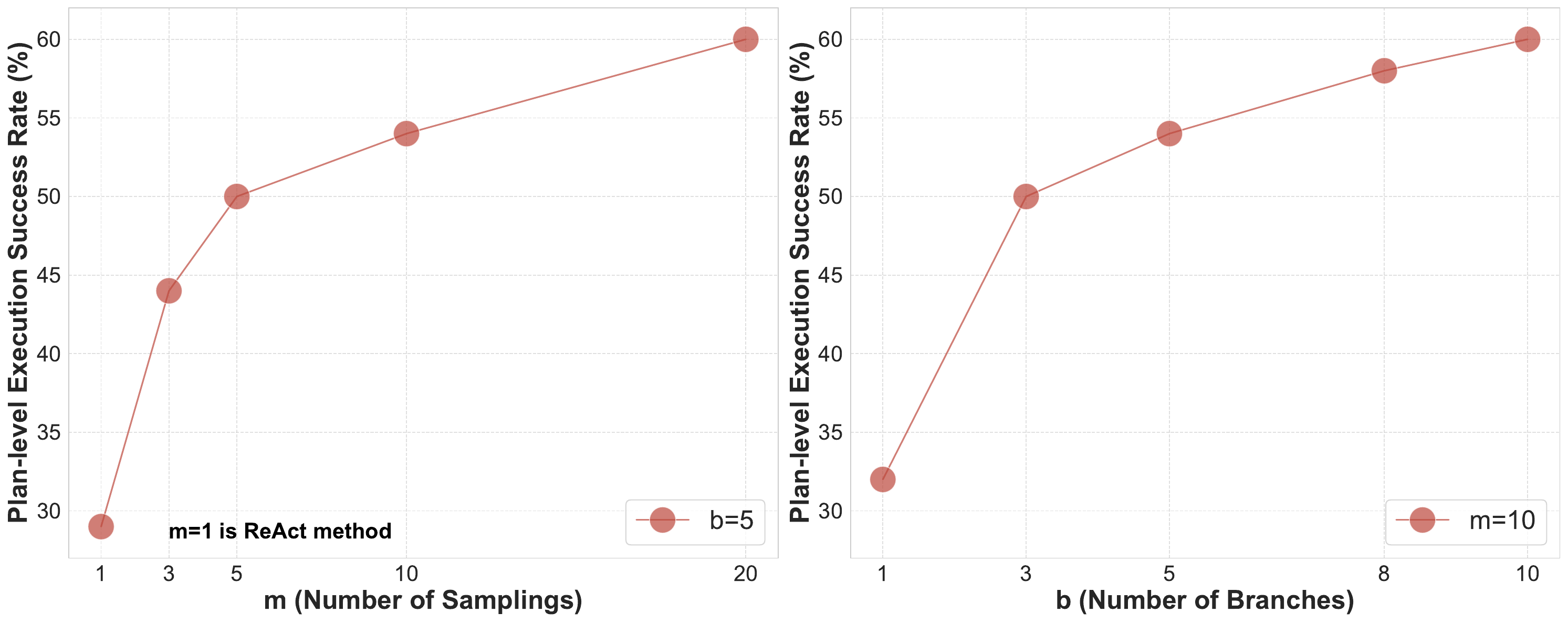}
    \caption{Evaluation of EGB search with Claude-Sonnet-4 on the \slate synthetic dataset with varied configurations: (a) number of samplings, $m$, used to estimate entropy, and (b) budget, $b$, as the maximum number of branch trials.}
    \vspace*{-0.2in}
    \label{fig:hyperparam_plot}
\end{figure}

\textbf{Impact of Hyperparameters:} We systematically investigate EGB's two critical hyperparameters, as illustrated in Figure 4. The number of samplings $m$ controls entropy estimation quality: performance increases from 29\% at $m=1$ to 60\% at $m=20$. Similarly, the branching budget $b$ governs the extent of error correction: success rate improves from 32\% at $b=1$ to 60\% at $b=10$. Both parameters exhibit diminishing returns beyond certain thresholds; thus, we adopt $m=10$ and $b=5$ as our default configuration to balance performance gains against computational cost. Notably, when $m=1$, EGB degenerates to the ReAct method without leveraging uncertainty signals. When $b=1$, no error-correction branches are triggered, so performance is determined solely by the initial trajectory quality. The key takeaway is that performance improves with both more accurate entropy estimation and increased branching opportunities for error correction.
\vspace{-1mm}
\section{Conclusion}
\label{sec:conclusion}
\vspace{-1mm}

In this work, we introduced \slate, a large scale benchmark for assessing tool augmented agents in long horizon e-commerce tasks. Our evaluation revealed that current methods struggle with efficiency and self-correction in vast action spaces. To address this, we proposed Entropy Guided Branching, an uncertainty aware search algorithm that dynamically allocates exploration effort where predictive entropy is high. Experimental results demonstrate that EGB significantly improves task success rates and computational efficiency. Together, \slate and EGB provide a robust foundation for building reliable LLM agents for complex environments. Additional details are provided in Appendix.

\section{Limitations}
Despite the robust performance of EGB and the comprehensive nature of \slate, several limitations warrant further investigation.

\textbf{Limitations of \slate.} While \slate incorporates intricate sequential dependencies and context-aware responses, it remains a synthetic benchmark. Its simulator is designed to return deterministic outcomes for matched tool invocations to enable controlled plan-level evaluation, which does not fully capture operational behaviors of real APIs such as latency spikes, retry policies, transient invocation failures, and external service instability. In addition, \slate currently evaluates tool-invocation decisions under a fixed reference plan, so it does not directly assess an agent's ability to synthesize high-quality plans from scratch. Finally, since both the benchmark and simulator were generated through a closed-source LLM pipeline (Claude 3.7), the dataset may contain model-specific biases. This can make cross-backbone comparisons less strictly fair; therefore, our main conclusions emphasize comparisons across tool-use paradigms under the same backbone LLM.

\textbf{Limitations of EGB.} Although EGB-Sampling is fully compatible with black-box models, its reliance on multiple independent samplings ($m=10$) to estimate entropy introduces higher inference latency and API costs compared to single-pass greedy methods, potentially limiting its application in real-time or resource-constrained scenarios. Deterministic replay in our experiments is a property of the \slate simulator and a practical implementation of backtracking, rather than a fundamental constraint of EGB. This assumption is also shared by tree-search baselines such as LATS, and practical systems can often support limited backtracking through checkpointing, idempotent API design, and rollback mechanisms; extending EGB to fully non-deterministic stateful environments remains future work. Moreover, this study focuses on e-commerce tasks; further work is needed to validate the generalizability of entropy-guided search in other tool-rich environments, such as software engineering or scientific research.

\section{Ethical Considerations}
This work focuses on improving the evaluation and planning efficiency of tool-augmented LLM agents through a synthetic benchmark and an uncertainty-aware search strategy. The proposed dataset does not contain real user data, personal information, or proprietary APIs, thereby minimizing privacy and data misuse risks. However, more efficient planning and execution may lower the barrier to deploying highly autonomous agents, which could be misused if applied without appropriate safeguards. In particular, aggressive exploration strategies may be undesirable in safety-critical or high-stakes environments. We emphasize that our methods are intended for controlled research settings and should be combined with external safety mechanisms, access controls, and human oversight in real-world deployment. Additionally, as the benchmark is synthetically generated, it may inherit biases from the underlying language model, and results should be interpreted with appropriate caution.


\bibliography{references}

\clearpage
\onecolumn

\begin{center}
{\Large\bfseries Appendices}
\end{center}
\vspace{0.5em}

\appendix

\section{Examples}
\label{app:examples}
This section presents a representative example from the \texttt{SLATE} dataset, illustrating the transformation from a user query to a hierarchical execution plan, associated tool calls, and deterministic simulation outputs. The example demonstrates the plan structure and execution semantics used for step-wise and plan-level evaluation.

\begin{figure*}[h]
\begin{promptwrapper}{Case}
\noindent
I want to create a limited-time summer promotion for my
\hilite{\textbf{``ThermoFlex Water Bottle''}} \;
(\textbf{SKU:} \incode{TF-WB-2023})
that offers \textbf{15\% off} when customers \textbf{buy 2 or more units}.
The promotion should run from \textbf{June 1, 2024} to \textbf{August 31, 2024},
with a minimum purchase requirement of \textbf{\$35.00}.
How do I set this up and create a promo code \hilite{\textbf{``SUMMERTF24''}}
that customers can use at checkout?
\end{promptwrapper}

\begin{promptwrapper}{Plan}
\begin{itemize}[leftmargin=1.2em,itemsep=2pt,topsep=2pt]
  \item \textcolor{grayText}{\textbf{step}}: \textcolor{grayText}{1. Verify product information}
  \begin{itemize}[leftmargin=1.4em,itemsep=1pt,topsep=1pt]
    \item \textcolor{grayText}{\textbf{step}}: \textcolor{grayText}{1.1 Retrieve product details}
    \item \textcolor{grayText}{\textbf{action}}: \keyw{get\_product\_details}\,(\textbf{sku}=\incode{``TF-WB-2023''})
  \end{itemize}

  \item \textcolor{grayText}{\textbf{step}}: \textcolor{grayText}{2. Create the promotion}
  \begin{itemize}[leftmargin=1.4em,itemsep=1pt,topsep=1pt]
    \item \textcolor{grayText}{\textbf{step}}: \textcolor{grayText}{2.1 Set up promotion parameters}
    \item \textcolor{grayText}{\textbf{action}}:
    \keyw{create\_promotion}\,(
      \textbf{product\_id}=\incode{OUTPUT\_FROM\_STEP\_1.1.product\_id},
      \textbf{discount\_percentage}=15,
      \textbf{min\_quantity}=2,
      \textbf{min\_purchase}=35.00,
      \textbf{start\_date}=\incode{``2024-06-01''},
      \textbf{end\_date}=\incode{``2024-08-31''}
    )
  \end{itemize}

  \item \textcolor{grayText}{\textbf{step}}: \textcolor{grayText}{3. Configure promo code}
  \begin{itemize}[leftmargin=1.4em,itemsep=1pt,topsep=1pt]
    \item \textcolor{grayText}{\textbf{step}}: \textcolor{grayText}{3.1 Create promo code}
    \item \textcolor{grayText}{\textbf{action}}:
    \keyw{create\_promo\_code}\,(
      \textbf{promotion\_id}=\incode{OUTPUT\_FROM\_STEP\_2.1.promotion\_id},
      \textbf{code}=\hilite{\texttt{``SUMMERTF24''}}
    )
  \end{itemize}

  \item \textcolor{grayText}{\textbf{step}}: \textcolor{grayText}{``4. Verify and activate the promotion''}
  \begin{itemize}[leftmargin=1.4em,itemsep=1pt,topsep=1pt]
    \item \textcolor{grayText}{\textbf{step}}: \textcolor{grayText}{``4.1 Validate promotion setup''}
    \item \textcolor{grayText}{\textbf{action}}:
    \keyw{validate\_promotion}\,(
      \textbf{promotion\_id}=\incode{OUTPUT\_FROM\_STEP\_2.1.promotion\_id}
    )
    \item \textcolor{grayText}{\textbf{step}}: \textcolor{grayText}{``4.2 Activate the promotion''}
    \item \textcolor{grayText}{\textbf{action}}:
    \keyw{activate\_promotion}\,(
      \textbf{promotion\_id}=\incode{OUTPUT\_FROM\_STEP\_2.1.promotion\_id},
      \textbf{promo\_code\_id}=\incode{OUTPUT\_FROM\_STEP\_3.1.promo\_code\_id}
    )
  \end{itemize}
\end{itemize}
\end{promptwrapper}
\caption{Example of plan generation and tool execution.}
\end{figure*}

\begin{figure*}[hp]

\begin{promptwrapper}{Simulate Outputs with context}
\begin{itemize}[leftmargin=1.2em,itemsep=3pt,topsep=2pt]
  \item \keyw{get\_product\_details}\,(\textbf{sku}=\incode{``TF-WB-2023''}) \\
  \textcolor{noteBlue}{$\Rightarrow$ Output:} \textbf{product\_id}=\incode{``P-TF-WB-2023-001''}

  \item \keyw{create\_promotion}\,(
    \textbf{product\_id}=\incode{``P-TF-WB-2023-001''},
    \textbf{discount\_percentage}=15,
    \textbf{min\_quantity}=2,
    \textbf{min\_purchase}=35.00,
    \textbf{start\_date}=\incode{``2024-06-01''},
    \textbf{end\_date}=\incode{``2024-08-31''}
  ) \\
  \textcolor{noteBlue}{$\Rightarrow$ Output:} \textbf{promotion\_id}=\incode{``PROMO-TF-2024-S001''}

  \item \keyw{create\_promo\_code}\,(
    \textbf{promotion\_id}=\incode{``PROMO-TF-2024-S001''},
    \textbf{code}=\hilite{\texttt{``SUMMERTF24''}}
  ) \\
  \textcolor{noteBlue}{$\Rightarrow$ Output:} \textbf{promo\_code\_id}=\incode{``PC-SUMMERTF24-001''}

  \item \keyw{validate\_promotion}\,(\textbf{promotion\_id}=\incode{``PROMO-TF-2024-S001''})

  \item \keyw{activate\_promotion}\,(
    \textbf{promotion\_id}=\incode{``PROMO-TF-2024-S001''},
    \textbf{promo\_code\_id}=\incode{``PC-SUMMERTF24-001''}
  ) \\
  \textcolor{noteBlue}{$\Rightarrow$ Output:} \textbf{success}=\incode{``true''}
\end{itemize}
\end{promptwrapper}

\begin{promptwrapper}{Deterministic Simulation Dictionary}
\noindent \textcolor{noteBlue}{\textbf{1.}} \keyw{get\_product\_details} \; \textcolor{grayText}{(Args: sku)} \\
\hspace*{1.2em}\incode{TF-WB-2023} \;\; \textcolor{noteBlue}{$\Rightarrow$}\;\; \incode{P-TF-WB-2023-001} \\
\hspace*{1.2em}\textcolor{grayText}{Other value} \;\; \textcolor{noteBlue}{$\Rightarrow$}\;\; \textcolor{grayText}{different default value}

\vspace{4pt}
\noindent \textcolor{noteBlue}{\textbf{2.}} \keyw{create\_promotion} \; \textcolor{grayText}{(Args: product\_id, discount\_percentage, min\_quantity, min\_purchase, start\_date, end\_date)} \\
\hspace*{1.2em}\incode{P-TF-WB-2023-001, 15, 2, 35, 2024-06-01, 2024-08-31}
\;\; \textcolor{noteBlue}{$\Rightarrow$}\;\; \incode{PROMO-TF-2024-S001} \\
\hspace*{1.2em}\textcolor{grayText}{Other value} \;\; \textcolor{noteBlue}{$\Rightarrow$}\;\; \textcolor{grayText}{different default value}

\vspace{4pt}
\noindent \textcolor{noteBlue}{\textbf{3.}} \keyw{create\_promo\_code} \; \textcolor{grayText}{(Args: promotion\_id, code)} \\
\hspace*{1.2em}\incode{PROMO-TF-2024-S001, SUMMERTF24}
\;\; \textcolor{noteBlue}{$\Rightarrow$}\;\; \incode{PC-SUMMERTF24-001} \\
\hspace*{1.2em}\textcolor{grayText}{Other value} \;\; \textcolor{noteBlue}{$\Rightarrow$}\;\; \textcolor{grayText}{different default value}
\end{promptwrapper}
\caption{Deterministic simulation dictionary for tool calls.}
\end{figure*}

\newpage
\clearpage

\section{Supplementary Experiments}
\label{app:experiments}

\subsection{Leveraging Past Executions with Memory}
\label{sec:app_memory}

\begin{table}[ht]
\caption{Evaluation of search methods with vs. without memory on the \slate synthetic dataset with Claude-4-Sonnet.}
\vspace{-0.1in}
\centering
\resizebox{0.6\linewidth}{!}{  
\begin{tabular}{c c c c}
\toprule
\multirow{4}{*}{\makecell{\textbf{Method}}} & \multicolumn{1}{c}{\textbf{Plan-level}} & \multicolumn{2}{c}{\textbf{Step-wise Evaluation}} \\
\cmidrule(lr){2-2} \cmidrule(lr){3-4}
 & \makecell{Execution\\Success\\Rate} & \makecell{Tool\\Match\\Rate} & \makecell{Action\\Identification\\Accuracy} \\
\midrule
\makecell{\textbf{ReAct}} & 29.3 $\pm$ 1.2 & 66.4 $\pm$ 0.4 & 85.7 $\pm$ 0.1 \\
\makecell{\textbf{Reflexion}} & 44.7 $\pm$ 2.3 & 61.5 $\pm$ 1.6 & 83.4 $\pm$ 0.5 \\
\makecell{\textbf{EGB-Sampling (Ours)}} & 54.0 $\pm$ 2.0 & 68.5 $\pm$ 1.5 & 87.9 $\pm$ 0.2 \\
\midrule
\makecell{\textbf{ReAct-with-Memory}} & 86.3 $\pm$ 0.6 & 92.7 $\pm$ 0.3 & 98.7 $\pm$ 0.1 \\
\makecell{\textbf{Reflexion-with-Memory}} & 91.0 $\pm$ 1.5 & 92.7 $\pm$ 1.1 & 98.4 $\pm$ 0.3 \\ 
\makecell{\textbf{EGB-with-memory (Ours)}} & \textbf{92.7 $\pm$ 1.0} & 92.6 $\pm$ 0.5 & 98.7 $\pm$ 0.2 \\
\bottomrule
\end{tabular}
}
\label{tab:performance_with_memory}
\end{table}

\textbf{Experiment Setup.} While EGB demonstrates strong performance through entropy-guided exploration, we investigate whether incorporating memory from previous task executions can further enhance its capabilities. We create a pseudo in-distribution setting by randomly partitioning the dataset $\Dcal$ into a validation set $\Dcal_\text{val}$ (50 samples) and a test set $\Dcal_\text{test}$ (50 samples), where both sets are drawn from the same task distribution but contain no shared tasks. We accumulate execution traces from the validation set as memory and evaluate on the test set.

\textbf{Analysis.} As shown in Table~\ref{tab:performance_with_memory}, memory substantially improves performance across all methods: ReAct improves from 29.3\% to 86.3\% (+57.0\%), Reflexion from 44.7\% to 91.0\% (+46.3\%), and EGB from 54.0\% to 92.7\% (+38.7\%). Notably, even with memory enabled, EGB maintains its advantage over baselines, outperforming ReAct-with-Memory by 6.4\% and Reflexion-with-Memory by 1.7\%. These results demonstrate that while memory provides a powerful mechanism for leveraging past experiences in similar tasks, EGB continues to offer systematic improvements regardless of whether memory is available, confirming its effectiveness as a complementary approach to memory-based methods.

\subsection{Error Entropy Correlation}
\label{sec:app_error_entropy}

\begin{table}[h]
\centering
\caption{Entropy-error relationship on SLATE using Qwen-logits.}
\label{tab:app_entropy_error}
\begin{tabular}{lccc}
\toprule
Entropy Range & \#Steps & \#Errors & Error Rate \\
\midrule
$[0.000, 0.640)$ & 365 & 184 & 50.7\% \\
$[0.640, 1.281)$ & 116 & 79 & 68.1\% \\
$[1.281, 1.921)$ & 67 & 52 & 77.6\% \\
$[1.921, 2.561)$ & 21 & 20 & 95.2\% \\
$[2.561, 3.202)$ & 15 & 14 & 93.3\% \\
\bottomrule
\end{tabular}
\end{table}

\textbf{Experiment Setup.} We use the Qwen-logits experiment results because they provide calibrated per-step logit distributions for entropy estimation. Entropy is computed at each tool-selection step, excluding fallback steps where logit computation fails. We compare steps whose first-pass selected tool matches the reference tool (\textit{Correct}, $n=233$) against those that do not (\textit{Error}, $n=349$), for a total of 582 analyzed steps. Error rate in Table~\ref{tab:app_entropy_error} is defined as the fraction of steps in each entropy bin whose first-pass tool does not match the reference.

\textbf{Analysis.} Correct steps have substantially lower entropy (mean $=0.052$, median $=0.000$) than error steps (mean $=0.119$, median $=0.081$), indicating stronger confidence when the tool choice is correct. Across bins, error rate rises from 50.7\% in the lowest-entropy range to above 93\% in the highest ranges, showing a near-monotonic trend. The lowest-entropy bin still has non-trivial error due to benchmark difficulty: with 1000 tools and functionally similar candidates, the model can be confidently wrong. Overall, entropy is best interpreted as a relative prioritization signal for branching rather than a binary error detector.

\textbf{How Branching Reduces Error.} We further analyze 68 successful branch corrections (68 out of 114 first-pass failed plans corrected by entropy-guided exploration). We identify three mechanisms: (1) \textbf{Direct fix} (17.6\%), where the branched step switches to the reference tool; (2) \textbf{Cascade fix} (30.9\%), where the branched step changes execution context and downstream tools become correct (on average, each branch changes 0.9 subsequent steps); and (3) \textbf{Functional equivalence} (51.5\%), where the branched tool name differs from the reference but yields equivalent outputs. These mechanisms explain why entropy-guided branching remains effective even when exact tool-name matching is not the only path to success.

\subsection{Cross-Benchmark Validation on Adapted UltraTool}
\label{sec:app_ultratool}

\begin{table}[htbp]
\centering
\caption{Evaluation of different methods on adapted UltraTool with GPT-OSS-120b.}
\label{tab:app_ultratool_results}
\begin{tabular}{lccc}
\toprule
\textbf{Methods} & \textbf{Execution Success Rate} & \textbf{Input Tokens} & \textbf{Output Tokens} \\
\midrule
DirectLLM        & 67.3 $\pm$ 4.7 & $2.65 \times 10^{4}$ & $7.17 \times 10^{2}$ \\
ReAct            & 69.0 $\pm$ 2.0 & $2.87 \times 10^{4}$ & $8.68 \times 10^{2}$ \\
Self-Consistency & 69.3 $\pm$ 1.3 & $2.87 \times 10^{5}$ & $8.65 \times 10^{3}$ \\
Reflexion        & 77.3 $\pm$ 3.7 & $1.56 \times 10^{5}$ & $4.85 \times 10^{3}$ \\
EGB-Sampling     & 79.7 $\pm$ 0.7 & $1.69 \times 10^{5}$ & $5.36 \times 10^{3}$ \\
EGB-Logits       & 82.0 $\pm$ 1.0 & $3.75 \times 10^{4}$ & $1.79 \times 10^{3}$ \\
LATS-10          & 71.3 $\pm$ 4.3 & $1.39 \times 10^{5}$ & $4.95 \times 10^{3}$ \\
LATS-20          & 74.0 $\pm$ 3.0 & $2.21 \times 10^{5}$ & $6.37 \times 10^{3}$ \\
LATS-30          & 79.7 $\pm$ 1.3 & $2.96 \times 10^{5}$ & $8.41 \times 10^{3}$ \\
LATS-40          & 80.3 $\pm$ 2.3 & $3.64 \times 10^{5}$ & $1.04 \times 10^{4}$ \\
LATS-50          & 82.0 $\pm$ 2.0 & $4.30 \times 10^{5}$ & $1.22 \times 10^{4}$ \\
\bottomrule
\end{tabular}
\end{table}

\textbf{Experiment Setup.} To reduce potential model-family circularity from SLATE (synthesized with Claude-family models), we perform external validation on UltraTool with GPT-OSS-120b. UltraTool does not fully match our setting (limited long-horizon structure, limited grounded execution outputs, and 1000-query test size), so we adapt it by constructing an executable library of 436 tools and evaluate the first 100 test queries. We include two additional baselines in this setting: Self-Consistency (10 end-to-end ReAct samples with plan-level majority vote) and LATS with budgets 10/20/30/40/50.

\textbf{Analysis.} Table~\ref{tab:app_ultratool_results} shows that EGB-Logits (82.0\%) matches LATS-50 (82.0\%) while using only a fraction of the token cost, and outperforms all other methods. EGB-Sampling (79.7\%) also achieves high performance, matching LATS-30 (79.7\%) with substantially fewer tokens. We also observe a clear budget--accuracy trade-off for LATS: success improves from 71.3\% (LATS-10) to 82.0\% (LATS-50), with diminishing returns as the search budget increases. These results support an accuracy--efficiency advantage for entropy-guided branching rather than universal dominance at all budgets.

\subsection{Robustness to Inaccurate Plan Specifications}
\label{sec:app_inaccurate_plan}

\begin{table}[htbp]
\centering
\caption{Execution success rate under accurate vs. inaccurate plans on adapted UltraTool with GPT-OSS-120b.}
\label{tab:app_plan_comparison}
\begin{tabular}{lcc}
\toprule
\textbf{Methods} & \textbf{Accurate Plan} & \textbf{Inaccurate Plan} \\
\midrule
ReAct            & 69.0 $\pm$ 2.0 & 65.0 $\pm$ 5.0 \\
Self-Consistency & 69.3 $\pm$ 1.3 & 65.3 $\pm$ 0.7 \\
EGB-Sampling     & 79.7 $\pm$ 0.7 & 72.0 $\pm$ 2.0 \\
\bottomrule
\end{tabular}
\end{table}

\textbf{Experiment Setup.} To test robustness to imprecise planning, we modify original UltraTool plans by rewriting some tool-call sub-steps into broader descriptions while preserving intent (e.g., ``Use file writing tool to create and write content'' $\rightarrow$ ``Perform file operation'', ``Search for files matching pattern'' $\rightarrow$ ``Find relevant files''). In total, 10.5\% of tool-call sub-steps are modified.

\textbf{Analysis.} All methods degrade when plan specificity drops, but EGB-Sampling remains strongest (72.0\%). Notably, EGB-Sampling under inaccurate plans still exceeds ReAct and Self-Consistency under accurate plans, indicating that entropy-guided branching provides robustness to upstream plan imprecision.

\newpage
\clearpage

\section{Algorithms}
\label{app:algorithms}
\subsection{Pseudocode}
\label{sec:app_pseudocode}
\small
\begin{algorithm}[!ht]
\caption{Entropy-Guided Branching (EGB)}
\label{alg:egb}
\begin{algorithmic}[1]
\REQUIRE Query $q$, Plan $p = \{ss_{1,1}, \ldots, ss_{n,m_n}\}$, Tool library $\mathcal{T}$, Simulator $\mathcal{S}$, Global branch limit $B=50$, Per-step branch limit $B_s=5$

\STATE \textbf{// Phase 1: Initial Execution with Entropy Recording}
\STATE $H \leftarrow \langle \rangle$ \COMMENT{Execution history}
\STATE $\mathcal{E} \leftarrow \langle \rangle$ \COMMENT{Entropy tree}

\FOR{each substep $ss_{i,j} \in p$}
    \STATE $\mathcal{C}_{i,j} \leftarrow \textsc{GetCandidates}(q, ss_{i,j}, H, \mathcal{T})$ \COMMENT{Retrieve candidate tools via embedding search}
    \STATE $(a_{i,j}^*, E_{i,j}, \mathcal{D}_{i,j}) \leftarrow \textsc{ComputeEntropy}(q, ss_{i,j}, H, \mathcal{C}_{i,j})$ \COMMENT{Alg.~\ref{alg:entropy-sampling} or \ref{alg:entropy-logits}}
    \STATE $o_{i,j} \leftarrow \mathcal{S}(a_{i,j}^*)$ \COMMENT{Execute selected action}
    \STATE $H \leftarrow H \cup \langle (ss_{i,j}, a_{i,j}^*, o_{i,j}) \rangle$
    \STATE $\mathcal{E} \leftarrow \mathcal{E} \cup \langle (i, j, E_{i,j}, \mathcal{D}_{i,j}) \rangle$ \COMMENT{Store entropy and distribution}
\ENDFOR

\IF{$\textsc{TaskSuccess}(H)$}
    \RETURN $H$
\ENDIF

\STATE \textbf{// Phase 2: Entropy-Guided Branching}
\STATE $\mathcal{E}_{\text{sorted}} \leftarrow \textsc{SortByEntropy}(\mathcal{E}, \text{descending})$
\STATE $b \leftarrow 0$ \COMMENT{Global branch counter}

\FOR{each $(i, j, E_{i,j}, \mathcal{D}_{i,j}) \in \mathcal{E}_{\text{sorted}}$}
    \STATE $t_{i,j}^* \leftarrow \textsc{ExtractToolName}(a_{i,j}^*)$ \COMMENT{Tool selected in first pass}
    \STATE $\mathcal{T}_{\text{alt}} \leftarrow \{t : (t, p_t) \in \mathcal{D}_{i,j}, t \neq t_{i,j}^*\}$ \COMMENT{Alternative tools from distribution}
    \STATE Sort $\mathcal{T}_{\text{alt}}$ by probability in descending order
    \STATE $b_s \leftarrow 0$ \COMMENT{Per-step branch counter}
    \FOR{each $t' \in \mathcal{T}_{\text{alt}}$}
        \IF{$b \geq B$ \OR $b_s \geq B_s$}
            \STATE \textbf{break} \COMMENT{Budget exhausted}
        \ENDIF
        \STATE $a_{i,j}' \leftarrow \textsc{GetOrGenerateCall}(t', q, ss_{i,j}, H_{<(i,j)}, \mathcal{D}_{i,j})$ \COMMENT{Use cached call or generate params$^\dagger$}
        \STATE $H' \leftarrow H_{<(i,j)} \cup \langle (ss_{i,j}, a_{i,j}', \mathcal{S}(a_{i,j}')) \rangle$ \COMMENT{Branch from step $(i,j)$}
        \FOR{each subsequent substep $ss_{r,s}$ where $(r,s) \succ (i,j)$}
            \STATE $a_{r,s} \leftarrow \textsc{SelectTool}(q, ss_{r,s}, H', \mathcal{T})$ \COMMENT{Single-pass selection}
            \STATE $H' \leftarrow H' \cup \langle (ss_{r,s}, a_{r,s}, \mathcal{S}(a_{r,s})) \rangle$
        \ENDFOR
        \IF{$\textsc{TaskSuccess}(H')$}
            \RETURN $H'$
        \ENDIF
        \STATE $b \leftarrow b + 1$; $b_s \leftarrow b_s + 1$
    \ENDFOR
\ENDFOR
\RETURN $H$ \COMMENT{Return first-pass result}
\end{algorithmic}
\vspace{0.05in}
\raggedright
\footnotesize
$^\dagger$ For Alg.~\ref{alg:entropy-sampling}, uses pre-sampled calls stored in $\mathcal{D}_{i,j}$. For Alg.~\ref{alg:entropy-logits}, generates parameters on-demand via LLM.
\end{algorithm}

\begin{algorithm}[!ht]
\caption{Entropy Computation via Sampling (EGB-Sampling)}
\label{alg:entropy-sampling}
\begin{algorithmic}[1]
\REQUIRE Query $q$, Substep $ss_{i,j}$, History $H_{i,j}$, Candidates $\mathcal{C}_{i,j}$, Sample count $m$
\ENSURE Selected action $a_{i,j}^*$, Entropy $E_{i,j}$, Action distribution $\mathcal{D}_{i,j}$

\STATE $\texttt{votes} \leftarrow \{\}$ \COMMENT{Vote counts per tool}
\STATE $\texttt{calls} \leftarrow \{\}$ \COMMENT{Sampled tool calls per tool}

\FOR{$k = 1$ \TO $m$}
    \STATE $a_{i,j}^{(k)} \sim \pi(a_{i,j} \mid H_{i,j}, \mathcal{C}_{i,j})$ \COMMENT{Sample complete action (tool + params) from LLM}
    \STATE $t^{(k)} \leftarrow \textsc{ExtractToolName}(a_{i,j}^{(k)})$
    \STATE $\texttt{votes}[t^{(k)}] \leftarrow \texttt{votes}[t^{(k)}] + 1$
    \STATE $\texttt{calls}[t^{(k)}] \leftarrow \texttt{calls}[t^{(k)}] \cup \{a_{i,j}^{(k)}\}$ \COMMENT{Store sampled call for potential branching}
\ENDFOR

\STATE \textbf{// Compute entropy from vote distribution}
\FOR{each tool $t \in \texttt{votes}$}
    \STATE $p_t \leftarrow \texttt{votes}[t] / m$
\ENDFOR
\STATE $E_{i,j} \leftarrow -\sum_{t} p_t \log p_t$

\STATE \textbf{// Select majority tool (use pre-sampled call)}
\STATE $t^* \leftarrow \arg\max_{t} \texttt{votes}[t]$
\STATE $a_{i,j}^* \leftarrow \textsc{SelectOne}(\texttt{calls}[t^*])$ \COMMENT{Use one of the sampled calls}
\STATE $\mathcal{D}_{i,j} \leftarrow \{(t, p_t, \texttt{calls}[t]) : t \in \texttt{votes}\}$ \COMMENT{Store distribution with pre-sampled calls}

\RETURN $(a_{i,j}^*, E_{i,j}, \mathcal{D}_{i,j})$
\end{algorithmic}
\end{algorithm}

\begin{algorithm}[!ht]
\caption{Entropy Computation via Logits (EGB-Logits): Qwen2.5 Adaptation}
\label{alg:entropy-logits}
\begin{algorithmic}[1]
\REQUIRE Query $q$, Substep $ss_{i,j}$, History $H_{i,j}$, Candidates $\mathcal{C}_{i,j} = \{c_0, \ldots, c_{K-1}\}$ with $K\leq100$, Threshold $\tau=0.01$

\STATE $\texttt{prompt} \leftarrow \textsc{BuildPrompt}(q, ss_{i,j}, H_{i,j}, \mathcal{C}_{i,j})$ \COMMENT{Ask LLM to output tool index $0$ to $K{-}1$}
\STATE $\mathbf{x} \leftarrow \textsc{Tokenize}(\texttt{prompt})$

\STATE \textbf{// Get first token distribution over digits 0-9}
\STATE $\mathbf{l}^{(1)} \leftarrow \textsc{ForwardPass}(\mathbf{x})[-1]$ \COMMENT{Logits at last position}
\STATE $\mathbf{p}^{(1)} \leftarrow \textsc{Softmax}(\mathbf{l}^{(1)}_{[\text{`0'}:\text{`9'}]})$ \COMMENT{Probability over digit tokens}

\STATE \textbf{// Compute conditional second token distribution}
\STATE $\mathbf{P} \leftarrow \mathbf{0}^K$ \COMMENT{Probability for each candidate}
\FOR{$d_1 = 0$ \TO $9$}
    \STATE $\mathbf{x}' \leftarrow \textsc{Concat}(\mathbf{x}, \textsc{Token}(d_1))$
    \STATE $\mathbf{l}^{(2)} \leftarrow \textsc{ForwardPass}(\mathbf{x}')[-1]$
    \STATE $\mathbf{p}^{(2)} \leftarrow \textsc{Softmax}(\mathbf{l}^{(2)}_{[\text{`0'}:\text{`9'}]})$
    \STATE $p_{\text{end}} \leftarrow 1 - \sum_{d_2=0}^{9} \mathbf{p}^{(2)}_{d_2}$ \COMMENT{Probability of non-digit token (sequence end)}
    
    \IF{$d_1 < K$}
        \STATE $\mathbf{P}[d_1] \leftarrow \mathbf{P}[d_1] + \mathbf{p}^{(1)}_{d_1} \cdot p_{\text{end}}$ \COMMENT{Single-digit index}
    \ENDIF
    \FOR{$d_2 = 0$ \TO $9$}
        \STATE $idx \leftarrow d_1 \times 10 + d_2$
        \IF{$10 \leq idx < K$}
            \STATE $\mathbf{P}[idx] \leftarrow \mathbf{P}[idx] + \mathbf{p}^{(1)}_{d_1} \cdot \mathbf{p}^{(2)}_{d_2}$ \COMMENT{Two-digit index}
        \ENDIF
    \ENDFOR
\ENDFOR
\STATE $\mathbf{P} \leftarrow \mathbf{P} / \sum_k \mathbf{P}[k]$ \COMMENT{Normalize}

\STATE \textbf{// Compute entropy over all candidates}
\STATE $E_{i,j} \leftarrow -\sum_{k=0}^{K-1} \mathbf{P}[k] \log \mathbf{P}[k]$

\STATE \textbf{// Filter candidates by probability threshold for branching}
\STATE $\mathcal{C}_{i,j}^{\text{filtered}} \leftarrow \{c_k : \mathbf{P}[k] \geq \tau\}$

\STATE \textbf{// Select highest probability tool and generate parameters}
\STATE $k^* \leftarrow \arg\max_{k} \mathbf{P}[k]$
\STATE $a_{i,j}^* \leftarrow \textsc{GenerateParams}(c_{k^*}, q, ss_{i,j}, H_{i,j})$ \COMMENT{Generate params only for selected tool}
\STATE $\mathcal{D}_{i,j} \leftarrow \{(c_k, \mathbf{P}[k]) : c_k \in \mathcal{C}_{i,j}^{\text{filtered}}\}$ \COMMENT{Params generated lazily during branching}

\RETURN $(a_{i,j}^*, E_{i,j}, \mathcal{D}_{i,j})$
\end{algorithmic}
\end{algorithm}

\clearpage
\subsection{Complexity Analysis}
\label{sec:app_complexity}
Let $n$ denote the number of plan substeps, $m$ the number of entropy samples per step in EGB-Sampling, and $b$ the number of executed branch attempts in Phase~2 (bounded by $B$). We summarize the dominant computational costs in terms of LLM-scale operations.

\textbf{Phase 1 (initial execution + entropy recording).} For EGB-Sampling, each step performs entropy estimation with $m$ sampled calls, yielding $O(nm)$ dominant LLM complexity. For EGB-Logits, entropy is derived from direct token-level probabilities (effectively $m{=}1$), reducing Phase~1 to $O(n)$ dominant LLM complexity.

\textbf{Phase 2 (entropy-guided branching).} Each branch reuses the prefix before the branch point and re-executes only the suffix with single-pass tool selection. In the worst case, each attempt may replay $O(n)$ remaining steps, so Phase~2 is $O(bn)$; with branch points roughly centered, the average replay length is about $n/2$, i.e., $O(bn)$ up to constants.

\textbf{Total complexity.} Combining both phases gives:
\begin{itemize}
    \item EGB-Sampling (black-box entropy): $O\!\left(n(m+b)\right)$,
    \item EGB-Logits (white-box entropy): $O\!\left(n + bn\right)=O\!\left(bn\right)$ when branching dominates.
\end{itemize}
These expressions match the complexity trend in the accompanying algorithm diagram: entropy sampling dominates Phase~1 for black-box models, while branch replay dominates when branch budgets are large.
\newpage
\clearpage

\section{Prompt Templates}
\label{app:prompt_template}

\begin{figure*}[ht]
\begin{promptwrapper}{Prompt for Generating Diverse E-Commerce Cases}

Generate ONE focused e-commerce query that an \{user\_type\} might have.

User type: \{user\_type\}
Focus area: \{focus\}
Complexity level: \{complexity\}

Requirements based on COMPLEXITY LEVEL:

\{complexity\} COMPLEXITY GUIDELINES:
\{self.\_get\_complexity\_guidelines(complexity)\}

For ALL queries, ensure:
\begin{enumerate}
\item Include SPECIFIC DETAILS that can be used as direct arguments in tools:
\begin{itemize}
\item Product names and identifiers (e.g., ``Sunset Yoga Mat (SKU: YM-2023-BL)'')
\item Order numbers (e.g., ``Order \#AB-12345678'')
\item FULL DATES WITH YEARS (e.g., ``January 15, 2023'' not just ``January 15'')
\item Prices with currency (e.g., ``\$49.99'')
\item Specific quantities (e.g., ``3 units'')
\end{itemize}
\item Include MULTIPLE DATA POINTS for tools to extract as parameters
\item For COMPLEX and ADVANCED queries, include:
\begin{itemize}
\item Multiple constraints or conditions
\item Timing requirements or deadlines
\item Preferences with priorities
\item Historical context or previous actions
\item Special exceptions or unusual circumstances
\end{itemize}
\end{enumerate}

Examples of queries at different complexity levels:

SIMPLE: ``I need to track my order \#RT-78256391 for the Samsung Galaxy Buds that I ordered on May 3, 2023. When will it be delivered?''

MODERATE: ``I need to return two items from my Order \#112-9384756 placed on March 12, 2023: the Samsung Galaxy S22 with a cracked screen and the protective case. I want to keep the screen protector and charging cable. Can I get a return label for just those two items?''

COMPLEX: ``I need to modify my bulk order \#BLK-2023-4872 placed on February 28, 2023 for my company. We ordered 50 Lenovo ThinkPad T14 laptops with i7 processors at \$1,299 each, but now I need to change 15 of them to the i9 model which costs \$1,599 each. We've already paid the deposit of \$25,000 and delivery is scheduled for June 15, 2023. I need to know if this change will affect our delivery date and what additional payment is required.''

ADVANCED: ``I'm managing our company's quarterly office supply order (PO \#BZ-45721) placed on April 2, 2023 with scheduled delivery on April 20, 2023 across three locations. For the Chicago office (Location ID: CHI-005), we need to cancel the 12 ergonomic chairs (\$259 each) due to their recent merger, but expedite the 15 monitor stands (\$89 each) to arrive by April 15. For the Boston office (Location ID: BOS-002), we need to add 8 wireless keyboards (\$65 each) and 8 wireless mice (\$45 each) for new hires. The New York office (Location ID: NYC-001) shipment is fine as is, but we need to change the delivery window to after 2:00 PM. We're eligible for the 12\% corporate discount and already applied the SPRING2023 promo code for 5\% off. How will these changes affect our total, and can all these modifications be accommodated before processing begins on April 10?''

Return only the query text with no additional explanation.

\end{promptwrapper}
\caption{Prompt to generate diverse e-commerce cases from different user perspectives. \textbf{Example Variable Values:} \texttt{user\_type} alternates between ``buyer'' and ``seller'' based on query ID; \texttt{focus} is randomly selected from buyer focus areas (``product search with multiple conflicting requirements'', ``order tracking for multiple items with shipping complications'', ``complex product return with partial refund request'', etc.) or seller focus areas (``updating complex product variations and attributes'', ``inventory management across multiple warehouses'', ``implementing tiered pricing strategy with conditions'', etc.); \texttt{complexity} is one of ``SIMPLE'', ``MODERATE'', ``COMPLEX'', or ``ADVANCED'' with weighted distribution of 0\%, 10\%, 80\%, 10\% respectively.}
\end{figure*}
\clearpage
\begin{figure*}[ht]
\begin{promptwrapper}{Prompt for Creating Specialized Tool Plans to Resolve E-Commerce Queries (Part 1)}

Create a plan with specialized tools to resolve this e-commerce query:

User Query (\{user\_type\}, \{complexity\} complexity): ``\{query\}''

\{existing\_tools\_text\}

Please provide:

1. A HIERARCHICAL PLAN with:
\{plan\_requirements\}

2. For each SUBSTEP, indicate:
\begin{itemize}
\item If it's a high-level step: ``tool'': ``null'' - NO EXCEPTIONS!
\item If no tool is needed: ``tool'': ``No tool required''
\item If a tool is needed: ``tool'': ``tool\_name(param1='value1', param2='value2')''
\end{itemize}

3. CRITICAL: FIRST TOOL CALL arguments must come DIRECTLY from the query
\begin{itemize}
\item Example: If query mentions ``Order \#AB-12345'', first tool should use order\_id='AB-12345'
\item Extract actual values from the query text, don't invent new values
\end{itemize}

4. CRITICAL: SUBSTEP DESCRIPTIONS MUST BE DISTINCT FROM TOOL NAMES
\begin{itemize}
\item Make each substep description MEANINGFUL and CONTEXT-RICH
\item Do not create tool names that are Identical or Very Similar to substep descriptions.
\item BAD: ``1.1 Get order details'' when using ``get\_order\_details'' tool
\item GOOD: ``1.1 Retrieve customer's purchase history for Order \#AB-123'' when using ``get\_order\_details'' tool
\item Describe the PURPOSE and CONTEXT of the step, not just the action
\item Include relevant business context and specific goals for each step
\end{itemize}

\end{promptwrapper}
\caption{Prompt for creating specialized tool plans (Part 1). \textbf{Example Variable Values:} \texttt{user\_type} is either ``buyer'' or ``seller''; \texttt{complexity} is one of ``SIMPLE'', ``MODERATE'', ``COMPLEX'', or ``ADVANCED''; \texttt{query} is the actual query text generated in stage 1. \textbf{existing\_tools\_text:} Dynamically generated list of top 30 most relevant tools from global tool library. \textbf{plan\_requirements:} SIMPLE (2-3 high-level steps, 1-2 substeps each, 3-5 total); MODERATE (3-4 high-level steps, 1-3 substeps each, 6-8 total); COMPLEX (4-5 high-level steps, 2-3 substeps each, 8-12 total); ADVANCED (5-7 high-level steps, 2-4 substeps each, 12-16 total).}
\end{figure*}

\begin{figure*}[ht]
\begin{promptwrapper}{Prompt for Creating Specialized Tool Plans to Resolve E-Commerce Queries (Part 2)}

5. TOOL SELECTION AND DESIGN:
\begin{itemize}
\item TRY REUSE EXISTING TOOLS whenever appropriate (from the list above)
\item If none of the existing tools are suitable, design NEW SPECIALIZED TOOLS that:
\begin{itemize}
\item Have SPECIFIC, FOCUSED functionality (not general-purpose)
\item Use SIMPLE arguments (2-3 parameters maximum)
\item Return SIMPLE, FOCUSED results (1-3 fields maximum)
\item Follow snake\_case naming convention
\item Avoid creating monolithic ``do everything'' tools
\end{itemize}
\end{itemize}

6. CRITICAL - AVOID TOOL REPETITION:
\begin{itemize}
\item DO NOT use the same tool in consecutive substeps (e.g., avoid ``2.1 xxx tool\_A; 2.2 xxx tool\_A'')
\item If you need to call the same tool multiple times, space them out with other operations
\item Each substep should ideally use a DIFFERENT tool to create diversity
\item Create specialized tools for different aspects rather than reusing generic ones
\end{itemize}

7. IMPORTANT - SEQUENTIAL DEPENDENCIES:
\begin{itemize}
\item Later steps should use results from previous steps for sequential dependencies
\item Use the format OUTPUT\_FROM\_STEP\_X.Y.field to reference previous outputs
\item Example: product\_id=OUTPUT\_FROM\_STEP\_1.2.product\_id
\item Create a CHAIN of dependencies where each step builds on previous results
\item Ensure that tool outputs from early steps provide necessary inputs for later steps
\end{itemize}

8. FINAL STEP should produce a DIRECT RESULT that resolves the query
\begin{itemize}
\item The last tool should return a clear outcome or answer
\item For example: confirmation message, success status, or direct result
\item FINAL STEPS should utilize outputs from earlier steps
\end{itemize}

9. FOR \{complexity\} COMPLEXITY:
\begin{itemize}
\item \{self.\_get\_plan\_complexity\_guidelines(complexity)\}
\end{itemize}

\end{promptwrapper}
\caption{Prompt for creating specialized tool plans (Part 2). Focus on tool selection, reuse strategy, avoiding repetition, and establishing sequential dependencies between steps.}
\end{figure*}

\begin{figure*}[h]
\begin{promptwrapper}{Prompt for Creating Specialized Tool Plans to Resolve E-Commerce Queries (Part 3)}

10. TOOL DIVERSITY REQUIREMENTS:
\begin{itemize}
\item Create tools that span different functional domains (retrieval, validation, processing, notification, etc.)
\item Avoid generic tools like ``process\_request'' or ``handle\_query''
\item Instead create specific tools like ``validate\_return\_window'', ``calculate\_refund\_amount'', ``send\_confirmation\_email''
\item Each tool should have a clear, single responsibility
\end{itemize}

\medskip
\textbf{TOOL DESIGN PRINCIPLES:}
\begin{enumerate}
\item FOCUSED: Each tool should do ONE thing well
\item COMPOSABLE: Tools should work together through their inputs/outputs
\item REUSABLE: Create tools that could be useful in other scenarios
\item SIMPLE: Prefer multiple simple tools over one complex tool
\item DIVERSE: Create tools spanning different functional domains
\item SEQUENTIAL: Design tools to create natural dependencies and data flow
\end{enumerate}

\medskip
\textbf{Remember:}
\begin{itemize}
\item REUSE existing tools whenever appropriate
\item FIRST TOOL must use arguments DIRECTLY from the query
\item Keep tools SPECIALIZED with SIMPLE inputs and outputs
\item Ensure steps are SEQUENTIAL with clear dependencies
\item AVOID repeating the same tool in consecutive steps
\item FINAL STEP should provide a DIRECT RESOLUTION to the query
\item Return valid JSON only
\end{itemize}

\end{promptwrapper}
\caption{Prompt for creating specialized tool plans (Part 3). Format response as JSON object with ``plan'' array (high-level steps with ``tool'': ``null'', and substeps with tool calls or ``No tool required'') and ``tools'' array (defining all tools with ``name'', ``description'', ``arguments'', and ``results'' fields). \textbf{Complexity Guidelines:} SIMPLE (1-2 tool calls, linear, minimal dependencies); MODERATE (2-4 tool calls, at least one dependency and conditional step); COMPLEX (4-7 tool calls, multiple dependencies, validation, branching paths); ADVANCED (7+ tool calls, multi-stage workflow, complex dependencies, edge case handling, maximum tool diversity).}
\end{figure*}
\clearpage
\begin{figure*}[ht]
\begin{promptwrapper}{Prompt for Generating Diverse E-Commerce Tools}

Generate \{batch\_size\} DIVERSE e-commerce tools with COMPLETELY UNIQUE NAMES.

\medskip
\textbf{TOOL NAMING REQUIREMENTS:}
\begin{enumerate}
\item Each tool name MUST be UNIQUE and use snake\_case format
\item NEVER use any of these existing names: \{names\_str\}
\item Create DISTINCTIVE names that avoid generic patterns
\item Consider these name patterns for inspiration: \{name\_pattern\_text\}
\item Each tool name should reflect its SPECIFIC FUNCTION and DOMAIN
\item VERIFY that each name is different from all others in your response
\end{enumerate}

\medskip
\textbf{TOOL ASSIGNMENTS} - Create exactly one tool for EACH of these specific domains:
\{domains\_text\}

\medskip
\textbf{SAMPLE TOOL STRUCTURE:}
\{json.dumps(sample\_tool, indent=2)\}

\medskip
\textbf{TOOL REQUIREMENTS:}
\begin{enumerate}
\item Match each tool precisely to its assigned domain above
\item Include 1-3 SIMPLE arguments with clear purposes
\item Return 1-3 FOCUSED result fields
\item Include ``query\_id'': ``dummy'' in each tool
\item Follow the exact JSON structure of the sample
\end{enumerate}

\medskip
Return a JSON array containing \{batch\_size\} tools with UNIQUE names:
\begin{verbatim}
[
  {tool1},
  {tool2},
  ...
]
\end{verbatim}

ONLY return the JSON array with no additional text.

\end{promptwrapper}
\caption{Prompt for generating diverse e-commerce tools with unique names. \textbf{Example Variable Values:} \texttt{batch\_size} is configurable (typically 10 or 20, default 10). \texttt{names\_str} is a comma-separated string of 20 random existing tool names to avoid duplication (e.g., ``get\_order\_details, validate\_return\_window, calculate\_refund, search\_products, update\_inventory, generate\_shipping\_label, process\_payment, send\_notification, track\_shipment, analyze\_sales, manage\_promotions, verify\_address, calculate\_tax, check\_stock, create\_invoice, update\_customer, generate\_report, schedule\_delivery, process\_return, validate\_coupon''). \texttt{name\_pattern\_text} consists of 4 randomly selected naming patterns from: ``feature\_specific\_domain, domain\_specific\_action, specialized\_task\_handler, domain\_analyzer, action\_target\_generator, domain\_insight\_provider, specialized\_workflow\_automation, target\_specific\_optimizer''. \texttt{domains\_text} is a numbered list of specific domains (one per tool) from 100 predefined diverse domains covering: Product management (catalog, variants, bundling, photography assessment, competitive analysis, recommendations, sourcing, limited editions, warranties, etc.), Shipping \& logistics (delivery, warehousing, returns, tracking, etc.), etc. \texttt{sample\_tool} is a randomly selected existing tool or default structure containing: \texttt{name}, \texttt{description}, \texttt{query\_id}, \texttt{arguments} (with type, properties, and argument details), and \texttt{results} (with type, properties, and result field details).}
\end{figure*}
\clearpage
\begin{figure*}[ht]
\begin{promptwrapper}{Prompt for Categorizing E-Commerce Tools into Functional Domains}

Analyze these e-commerce tools and categorize them into functional domains.

\medskip
\textbf{TOOLS TO CATEGORIZE:}
\{json.dumps(batch\_tools, indent=2)\}

\medskip
\textbf{PREDEFINED CATEGORIES:}
\{json.dumps(self.categories, indent=2)\}

\medskip
\textbf{INSTRUCTIONS:}
\begin{enumerate}
\item For each tool, assign it to the MOST APPROPRIATE category from the predefined list
\item If a tool could fit multiple categories, choose the PRIMARY function
\item Consider the tool's main purpose and typical use case
\item Return a JSON object mapping tool names to their categories
\end{enumerate}

\medskip
\textbf{Return format:}
\begin{verbatim}
{
  "tool_name_1": "Category Name",
  "tool_name_2": "Category Name",
  ...
}
\end{verbatim}

ONLY return the JSON object with no additional text.

\end{promptwrapper}
\caption{Prompt for categorizing e-commerce tools into functional domains. \textbf{Example Variable Values:} \texttt{batch\_tools} is a batch of 10-20 tools to categorize, each containing \texttt{name}, \texttt{description}, \texttt{arguments}, and \texttt{results} fields (e.g., ``get\_order\_details'', ``calculate\_shipping\_cost'', etc.). \texttt{self.categories} is a predefined list of 15 functional domain categories: ``Order Management'', ``Product Management'', ``Inventory Management'', ``Shipping \& Logistics'', ``Payment Processing'', ``Customer Management'', ``Returns \& Refunds'', ``Pricing \& Promotions'', ``Analytics \& Reporting'', ``Search \& Discovery'', ``Reviews \& Ratings'', ``Notifications \& Communication'', ``Authentication \& Security'', ``Marketplace Integration'', and ``Content Management''.}
\end{figure*}
\clearpage
\begin{figure*}[ht]
\begin{promptwrapper}{Prompt for Generating Tool Simulation Outputs}

Generate a realistic JSON output for this e-commerce tool call.

QUERY:
``\{context['query']\}''

CURRENT STEP:
``\{context['step']\}''

TOOL:
\begin{itemize}
\item Name: \{tool\_name\}
\item Description: \{tool\_def.get('description', 'No description available')\}
\end{itemize}

ARGUMENTS USED:
\{json.dumps(args, indent=2)\}

PREVIOUS TOOL OUTPUTS:
\{json.dumps(context['previous\_outputs'], indent=2) if context['previous\_outputs'] else ``No previous outputs''\}

REQUIRED OUTPUT PROPERTIES:
\{json.dumps(result\_props, indent=2)\}

EXAMPLE FORMAT:
\{json.dumps(example\_output, indent=2)\}

COMPLEXITY LEVEL: \{complexity\}
\{complexity\_guidance\}

INSTRUCTIONS:
\begin{enumerate}
\item Generate SPECIFIC, REALISTIC values consistent with the original query
\item Maintain CONSISTENCY with previous tool outputs
\item Include ALL required properties in the tool's result schema
\item Values should match their expected data types
\item \{``This is the FINAL TOOL in the plan. Make sure the output provides a DIRECT RESULT that clearly resolves the user's request (e.g., confirmation, status, or answer)'' if is\_final\_step else ``Keep output focused and relevant to the step''\}
\end{enumerate}

RETURN ONLY THE JSON OUTPUT OBJECT.

\end{promptwrapper}
\caption{Prompt for generating tool simulation outputs. \textbf{Example Variable Values:} \texttt{context['query']} is the original e-commerce query text; \texttt{context['step']} is the current step description from the plan; \texttt{tool\_name} is the name of the tool being called; \texttt{tool\_def} is the tool definition object containing description and other metadata; \texttt{args} is a JSON object containing the arguments passed to the tool; \texttt{context['previous\_outputs']} is a JSON object or array containing outputs from previously executed tools (null or empty if no previous outputs); \texttt{result\_props} is a JSON schema object defining required output properties with types and descriptions; \texttt{example\_output} is a sample JSON object showing the expected output format; \texttt{complexity} is one of ``SIMPLE'', ``MODERATE'', ``COMPLEX'', or ``ADVANCED''; \texttt{complexity\_guidance} is text providing specific guidance for generating outputs at the given complexity level; \texttt{is\_final\_step} is a boolean indicating whether this tool is the final step in the plan (affects instruction 5).}
\end{figure*}
\clearpage


\end{document}